\documentclass[11pt,a4paper]{article}
\usepackage[T1]{fontenc}
\usepackage[utf8]{inputenc}
\usepackage{lmodern}
\usepackage[margin=24mm]{geometry}
\usepackage{graphicx}
\usepackage{array,longtable}
\usepackage[table]{xcolor}
\usepackage{microtype}
\usepackage{xurl}
\usepackage[hidelinks]{hyperref}
\usepackage{placeins}
\usepackage{caption}
\definecolor{TableHead}{HTML}{E8EFF4}
\definecolor{TablePale}{HTML}{F8FAFB}
\definecolor{TableGrid}{HTML}{D9D9D9}
\hypersetup{pdftitle={DelegateEval: Evaluating Platforms for Personal External Service Agents},pdfauthor={Jichao Li, Jinrong Lu, Manjiang Yu, Zhixuan Liu}}
\title{DelegateEval: Evaluating Platforms for Personal External Service Agents}
\author{Jichao Li, Jinrong Lu, Manjiang Yu, Zhixuan Liu\\[4pt]\small BRICS Institute of Future Networks (China Branch)\\[3pt]\footnotesize\texttt{\{lijichao, lujinrong, yumanjiang, liuzhixuan\}@bifnc.cn}}
\date{}
\begin{document}
\maketitle
\begin{abstract}
AI agents are expanding from personal assistance into professional services delivered to external clients on an individual's behalf. We define a personal external service agent as one authorized and managed on an ongoing basis by an individual. It uses that person's knowledge, experience, and professional methods to accept client work, deliver results, and provide follow-up support. Client feedback, professional reputation, and potential income may give individuals more direct incentives to keep developing and improving agents for external services than for personal assistance. Yet building a safe, reliable external service system independently remains difficult, even with AI-assisted programming. People without AI expertise need platforms that are easy to use and safe to operate, with support for creating, developing, and running their agents. We propose the DelegateEval evaluation framework and experimental design to assess these platforms and guide their development. The framework covers eight dimensions: ease of creation and management, service discovery and matching, delivery quality, expertise inheritance, service continuity, learning and improvement, reliability and governance, and efficiency and value. A complete dataset, executable environment, and unified scoring protocol remain to be developed, and empirical validation across foundation models has yet to be conducted.
\end{abstract}

\FloatBarrier
\section{Introduction}\label{sec:1}
Personal agents are moving beyond one-off question answering toward systems that operate over extended periods and handle ongoing work. Meta's Muse, released in September 2026, provides one example. According to Meta's official description, Muse has a dedicated virtual machine and browser, can continue working after the user closes the application, retains relevant memories, and requests confirmation for sensitive actions \cite{ref1}. Together, these features support sustained operation by personal assistants. They also raise a further question: can an agent that works for an individual over time also represent that individual in providing services to external clients?

Personal external service agents offer a different way to organize professional services: an individual's accumulated knowledge, experience, and working methods can support more clients even when that person cannot participate in every task. Through appropriate configuration and development, an agent may apply an engineer's refined code and debugging methods, or a teacher's established ways of explaining concepts, to client work. We call an agent a personal external service agent when it is authorized and continually managed by a specific individual, draws on that person's professional resources, and accepts and fulfills commissions for external clients over time. A client may submit a request directly or delegate it through a personal assistant. This service model raises a central evaluation question: how can a platform demonstrate that an agent has inherited a particular person's expertise and can continue to honor its commitments to clients when that individual has only limited involvement?

Evaluating these platforms requires viewing each task as part of an ongoing service relationship. After an engineer's agent delivers working code, it may still need to maintain earlier versions, handle changing requirements, or respond to revoked authorization. Receipt of a file does not necessarily end the agreed service. A platform must help individuals from different backgrounds establish services, help clients find suitable providers, and support the use of professional resources, continuity of prior work, and accumulation of service experience. Evaluation must therefore track whether the individual's methods reduce professional errors, whether new requirements are handled without neglecting earlier commitments, and whether feedback improves subsequent tasks. It must also distinguish the contributions of the foundation model, the individual's resources, platform mechanisms, and additional human assistance, while recording all effort and resources used. These observations are needed to determine whether an initial successful delivery can lead to sustained service and whether quality, safety, and cost remain acceptable as service volume grows.

We therefore propose DelegateEval, a platform-level evaluation framework organized around agent creation, individual service episodes, and longitudinal service. It assesses eight dimensions: ease of creation and management, service discovery and matching, delivery quality, expertise inheritance, service continuity, learning and improvement, reliability and governance, and efficiency and value. Using predefined task requirements and independent acceptance evidence, the framework reports service performance, deliveries that meet the acceptance criteria, safety incidents, human effort, and total cost separately. It also records technical configurations and operating conditions to support comparisons across foundation models and implementation approaches. Expertise inheritance tests how effectively a fixed system version uses an individual's professional resources; learning and improvement tests the additional benefit of updates based on service experience that the system is authorized to use. The framework and evaluation design presented here provide a basis for future datasets, execution environments, common protocols, and cross-model experiments. The central question is whether, as services grow, a platform can bring an individual's expertise to more clients while sustaining an appropriate balance among that person's effort, delivery quality, and service responsibilities.

\FloatBarrier
\subsection{Main contributions}\label{sec:1.1}
This paper makes three contributions. First, it distinguishes personal assistants from personal external service agents by their service relationships and clarifies the responsibilities of the individual principal, client, agent, and platform. Second, it defines eight dimensions for evaluating how effectively platforms support professional services, treating service discovery, service continuity, and learning and improvement as separate measurement targets. A technical and operational configuration record documents the implementation conditions so that different technical approaches can be tested against the same service objectives. Third, it proposes comparisons across principals, tasks, and time, extending evaluation from individual tasks to long-term services and changes in service scale. These comparisons jointly consider delivery, fulfillment of commitments, learning, the principal's effort, economic returns, and risk control.

\FloatBarrier
\subsection{Organization of the paper}\label{sec:1.2}
Section~\ref{sec:2} reviews related work. Section~\ref{sec:3} defines the research scope and service relationships, and Section~\ref{sec:4} discusses the value of external services. Section~\ref{sec:5} presents the evaluation target and overall design; Section~\ref{sec:6} describes the eight dimensions of platform evaluation. Section~\ref{sec:7} discusses limitations and future work. The paper does not yet include an experimental protocol, scoring results, or appendices.

\FloatBarrier
\section{Related work and positioning}\label{sec:2}
\FloatBarrier
\subsection{Task completion and delivery}\label{sec:2.1}
SWE-bench evaluates code changes against real repository issues \cite{ref2}; AppWorld checks application states to determine whether a task has been completed and whether unintended changes have occurred \cite{ref3}; and TheAgentCompany evaluates file handling, software use, and collaboration with coworkers in a simulated organizational setting \cite{ref4}. \ensuremath{\tau}-bench combines multi-turn interaction, tool use, business rules, and final states, and examines consistency across repeated runs \cite{ref5}. Together, these benchmarks shift agent evaluation from the answers generated to the work actually completed. They also provide a methodological basis for our use of concrete deliverables and independent acceptance checks.

Intelligent AI Delegation further emphasizes verifiability in task decomposition. Its concept of "contract-first decomposition" requires task granularity to match the available verification methods: if a subtask produces a result that is too subjective, complex, or difficult to verify reliably, it should be decomposed recursively until completion can be established through automated tests, formal proofs, or another suitable verification mechanism \cite{ref6}. Task decomposition must therefore consider not only how work will be done but also, in advance, how its completion will be demonstrated.

DelegateEval extends this approach to the platform level. It examines whether individual tasks yield independently verifiable results, whether different principals can establish their own services, and whether their agents continue to provide verifiable results and honor commitments through subsequent tasks, feedback, and environmental changes.

\FloatBarrier
\subsection{Memory, learning, and social interaction}\label{sec:2.2}
Long-term service requires agents to use past information appropriately and improve through new experience. LongMemEval studies long-term memory capabilities such as information extraction, multi-session reasoning, temporal reasoning, and knowledge updates \cite{ref7}. LoRA provides parameter-efficient adaptation \cite{ref8}, while DPO supports preference learning \cite{ref9}. Both offer a technical basis for continued model adjustment. SOTOPIA studies social interaction among agents \cite{ref10}, Magentic Marketplace examines market behavior between consumer and merchant agents \cite{ref11}, and Intelligent AI Delegation discusses authorization, responsibility, and role boundaries in delegation relationships \cite{ref6}. These studies inform ongoing services from the perspectives of memory, learning, and collaboration among multiple parties.

Trainable add-on parametric memory offers another route to knowledge adaptation with a frozen foundation model. Wei et al. propose Deep Fusion Memory (DFM), which uses embeddings of text chunks obtained through offline retrieval as supervision for a memory predictor, then integrates the predicted memories into intermediate hidden states of the foundation model through gated cross-attention. The foundation model remains frozen during training while the new memory prediction and fusion modules are optimized jointly; at inference time, the system no longer explicitly retrieves from an external corpus \cite{ref12}. Here, "add-on" refers to a trainable module integrated into the model's internal computation, rather than a separate small model that answers questions independently.

SocialReasoning-Bench further distinguishes final outcomes from the quality of decisions made along the way. In tasks such as scheduling and negotiation, it assesses not only the outcome but also whether the agent made adequate use of available information and explored more appropriate options \cite{ref13}. Completing a task does not by itself establish that the delegated work was handled well. Personal external service agents face an additional requirement: they must continue to represent a particular principal while managing prior agreements, subsequent feedback, and changing client relationships. DelegateEval examines this distinction across tasks and over time, and compares performance across principals to assess whether a single platform can consistently support their different professional requirements.

\FloatBarrier
\subsection{Safety, interoperability, and platform-level challenges}\label{sec:2.3}
AgentDojo provides an executable test framework for studying attacks and defenses in environments with untrusted inputs, including prompt injection \cite{ref14}. For agent collaboration, A2A defines how agents exchange tasks, status information, and artifacts, supporting discovery of agent capabilities, task delegation, and artifact exchange \cite{ref15}. These efforts address safety and collaboration, respectively, in agent operation.

DelegateEval addresses a broader platform-level question: can one platform consistently help different individuals translate their professional requirements into services that external clients can discover, commission, receive, and have maintained over time? Safety mechanisms and interoperability protocols are important conditions for providing professional services, but their effectiveness must ultimately be assessed through actual service outcomes.

\FloatBarrier
\section{Research scope and service relationships}\label{sec:3}
\FloatBarrier
\subsection{Principals, external clients, and personal external service agents}\label{sec:3.1}
A personal external service agent is authorized by a specific individual to provide ongoing services to external clients using that person's knowledge, experience, preferences, and methods, within the service boundaries that person sets.

Here, principal refers only to the specific individual being represented. A personal external service agent represents that individual alone, rather than the individual's employer or another organization.

A personal assistant "does things for me"; a personal external service agent "provides services to others on my behalf." A programmer's personal assistant helps the programmer write code for their own work. The programmer's personal external service agent, by contrast, responds to external clients, accepts development commissions, and delivers work on the programmer's behalf. The service may be publicly accessible or available only to verified existing clients.

We use p for the represented individual (the principal), c for an external client, and \(a_p\) for that individual's personal external service agent. The principal determines the service scope, professional methods, knowledge sources, data boundaries, and management rules. The client proposes tasks, supplies materials, confirms requirements, and accepts deliverables. The agent acts on the requirements each party is authorized to set.

The external client must be distinguished from the party interacting directly with the service agent. Client c remains the individual or organization requesting the service, but may authorize a personal assistant or another client-side agent to discover services, submit commissions, communicate requirements, and receive deliverables. The personal external service agent may therefore interact with an agent acting for the client rather than with a human. This does not alter the underlying service relationship: the beneficiary, the parties represented on both sides, and their responsibilities must still be clear. A client-side agent may confirm requirements or accept deliverables only within the client's authorization.

Clients may adjust the requirements for the current delivery, while the principal controls the global service rules. If the principal changes the service scope, already accepted tasks must still be addressed. Client materials and the principal's private knowledge are managed separately, with permission obtained for each intended use: the current task, retention across tasks, or model training. When the parties' requirements conflict, the agent should seek clarification, negotiate, or request a decision, and record how existing commitments will be fulfilled or changed.

The personal external service agent \(a_p\) is a service role with explicit authorization and limits on action. It is not a simulation of the individual's voice or personality, nor does it imply granting the agent an independent legal personality. Professional licensing and liability must be assessed separately where applicable.

\begin{figure}[!htbp]
\centering
\includegraphics[width=\linewidth,height=0.60\textheight,keepaspectratio]{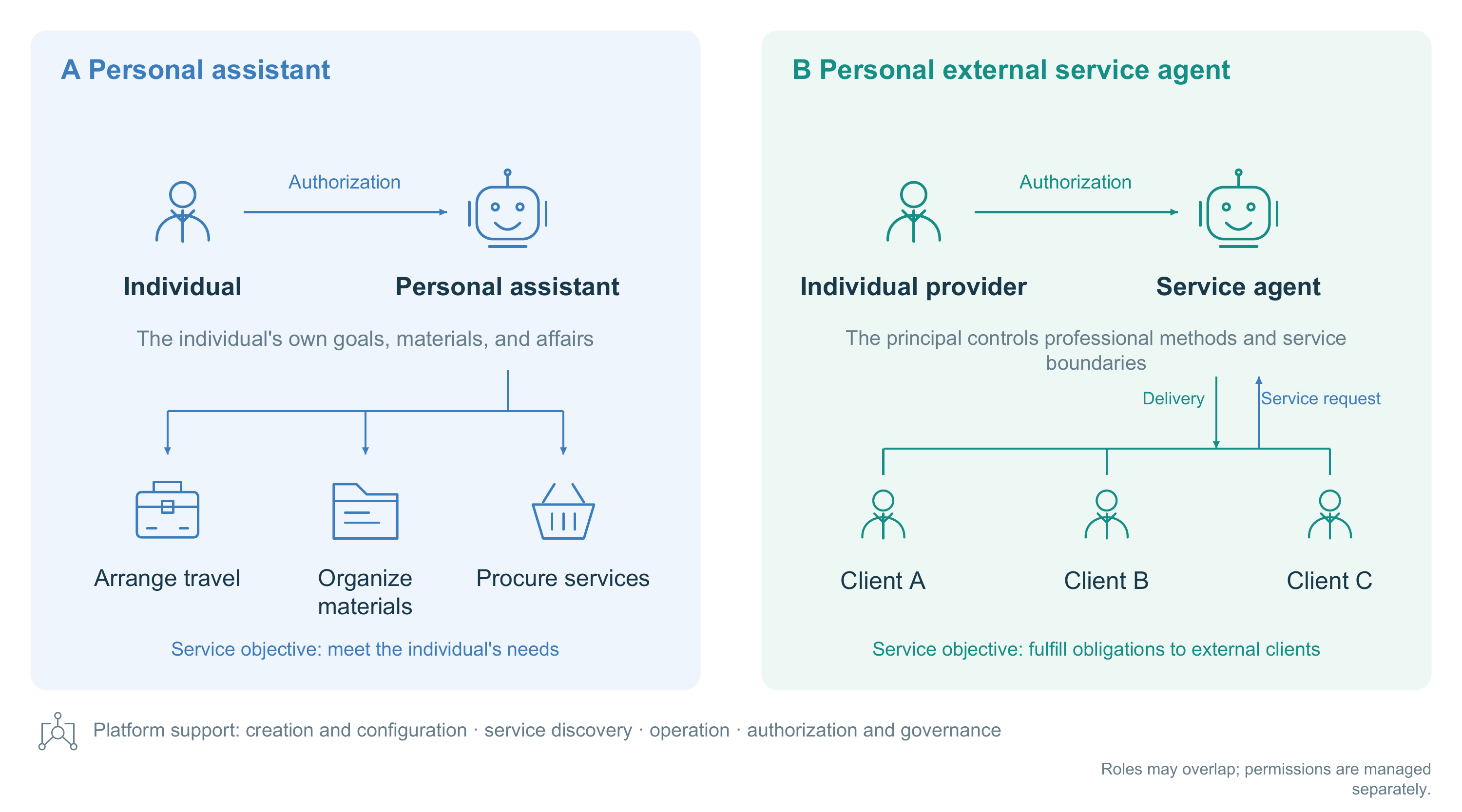}
\caption{Service relationships of personal assistants and personal external service agents. A personal assistant serves the individual's own needs. A personal external service agent is authorized and continually managed by a specific individual to accept work, deliver results, and provide maintenance for multiple external clients. One system may perform both roles, but identities, materials, permissions, and records are managed separately.}
\label{fig:1}
\end{figure}

Figure~\ref{fig:1} places the two service relationships side by side, distinguishing their service objectives and whom the agent represents.

\FloatBarrier
\subsection{Service roles and responsibility boundaries}\label{sec:3.2}
Writing code, editing documents, and editing videos describe what an agent can do. Personal assistance and personal external service describe whom it represents and serves. Task capabilities and service roles can be combined, and one system can assume different roles in different settings. Whether an agent falls within this paper's scope therefore depends on its actual service relationship.

A personal assistant can interact with external parties while still serving the individual's own needs. Booking a hotel, requesting a quote from a merchant, or sending an email on the individual's behalf are examples. The external service relationship considered here arises when an external client submits a request and an individual provider's agent accepts the work, delivers results, and undertakes the agreed follow-up service on that provider's behalf. Persistent availability, cloud deployment, and the ability to send messages are operational characteristics; none determines the service role on its own. This paper focuses on agents representing individuals. Agents that provide customer support, sales, or other services on behalf of companies or merchants represent organizations and operate under different responsibility arrangements. They fall outside the scope of this evaluation.

\begingroup\small
\arrayrulecolor{TableGrid}\setlength{\arrayrulewidth}{0.4pt}
\setlength{\tabcolsep}{6pt}\setlength{\extrarowheight}{3pt}\renewcommand{\arraystretch}{1.12}
\begin{longtable}{|>{\raggedright\arraybackslash}p{\dimexpr 0.24\linewidth-2\tabcolsep-1.3333333333333333\arrayrulewidth\relax}|>{\raggedright\arraybackslash}p{\dimexpr 0.29\linewidth-2\tabcolsep-1.3333333333333333\arrayrulewidth\relax}|>{\raggedright\arraybackslash}p{\dimexpr 0.47\linewidth-2\tabcolsep-1.3333333333333333\arrayrulewidth\relax}|}
\caption{Task organization and service roles. Task-oriented agents can perform different service roles, and one system may act as both a personal assistant and a personal external service agent.}\label{tab:1}\\
\hline
\rowcolor{TableHead} \textbf{Type or role} & \textbf{Identifying characteristics} & \textbf{Relation to this study} \\ \hline
\endfirsthead
\multicolumn{3}{l}{\small Table 1 (continued)}\\
\hline
\rowcolor{TableHead} \textbf{Type or role} & \textbf{Identifying characteristics} & \textbf{Relation to this study} \\ \hline
\endhead
\endfoot
Task-oriented agent & Execution organized around a specific task & May serve the individual or external clients; the actual service relationship determines the role \\ \hline
\rowcolor{TablePale} Personal assistant & Work organized around the individual's own needs & May find, purchase, and use external services on the individual's behalf, acting on the client side \\ \hline
Personal external service agent & Accepts external client work on behalf of a specific individual & The focus of this paper; responsible for delivery and follow-up within the agreed scope \\ \hline
\rowcolor{TablePale} Organizational service agent & Acts on behalf of a company, merchant, or other organization & Includes organizational customer support and sales roles; outside this paper's evaluation scope \\ \hline
\end{longtable}
\endgroup

When a system performs multiple roles, it must distinguish the sources and scope of instructions and manage identities, materials, permissions, and records separately. For example, a software engineer's client may request changes to the current deliverable, but cannot authorize the agent to read the engineer's private email, disclose another client's code, or permanently rewrite global development rules. The engineer may change the scope of future services, but delivery requirements already confirmed with a client must still be met or changed by agreement.

These boundaries should be tested in verifiable evaluation scenarios. Tests should examine whether the agent can identify whom it currently represents and serves, and what the requesting party is entitled to decide. When switching roles or handling work for different clients, tests should also check for inappropriate mixing of materials, use of incorrect permissions, and missed commitments.

\FloatBarrier
\subsection{Platform definition and evaluation boundaries}\label{sec:3.3}
A platform for personal external service agents enables different individuals to create, integrate, configure, run, publish, and manage their own agents, and gives external clients access to services and a way to commission work. A platform must first make three things clear: whom an agent represents, what services it can provide, and how a client can submit a request and receive the result. Other capabilities, such as ongoing agent development, public search, payment, collaboration with agents on other platforms, and local deployment, may depend on the service setting and platform's intended role; they are not mandatory for every platform.

The platform operator provides infrastructure and sets platform rules, but this does not change the boundaries on data use. Using the principal's or client's data for model training, long-term retention, or other purposes still requires the appropriate permission. Human assistance supplied by the platform for configuration, approval, expert review, or operations must also be recorded accurately and included in service costs, rather than being conflated with work completed independently by the agent.

We distinguish three levels: the model, the personal external service agent, and the platform. A model supplies basic capabilities such as reasoning, generation, and decision-making. A personal external service agent combines models with a specific individual's knowledge, experience, methods, service boundaries, and relevant tools to deliver a particular service to external clients. The platform supports different individuals in creating, running, and continually managing those services. DelegateEval therefore centers on the platform's ability to help different people reliably establish and sustain such services, rather than the strength of a particular model or agent alone. Evaluation evidence covers the agent creation process, actual service delivery, and longitudinal service records spanning tasks and time. Systems offering only a particular function or limited capability may also be evaluated on the relevant components, but conclusions remain limited to the scope tested.

\FloatBarrier
\section{The value of personal external service agents}\label{sec:4}
Personal external service agents combine an individual's distinctive expertise, AI-based task execution, and clearly defined service responsibilities to offer clients varied, accessible professional services with verifiable results and continuing accountability. Individuals can augment general-purpose models with their knowledge, experience, and working methods: they can build retrieval-based knowledge bases from proprietary materials they are authorized to use, organize professional judgment and operating methods into reusable skills and workflows, and combine general-purpose large models, smaller domain-specific models, and specialized tools as needed. When model access and training resources permit, they can also use fine-tuning or other post-training methods to adapt the parameters of models or add-on modules, and update knowledge, workflows, and model configurations in response to later service feedback. Even with the same foundation model, different individuals' agents can thus differ materially in their evidence base, judgment, execution, and delivery standards, offering specialized services that differ from those of standardized general-purpose AI tools.

For clients, using a personal external service agent instead of tools such as ChatGPT or Codex directly can reduce the burden of prompt design, task decomposition, tool selection, and repeated trial and error. Within the agreed scope, the agent and the provider behind it are responsible for applying professional judgment, verifying results, and making subsequent revisions. A client without domain expertise may receive an answer without being able to judge whether it is usable: working code may have broken existing functionality, a report full of charts may use an inappropriate analytical basis, and a complete mathematical solution may not mean that the student understands it. The service agent must select appropriate professional methods, check its work, and make necessary revisions to turn model-generated content into deliverables that meet the agreement and can pass acceptance checks.

This process uses a provider-side human-in-the-loop arrangement: the agent handles repeatable, standardizable work, while the individual provider conducts expert review, makes key decisions, and takes over when needed, remaining responsible for agreed follow-up service. Platforms should record which deliverables received only automated checks, which underwent human spot checks, which were confirmed by the principal, and the effort and resources used at each stage. Representing an expert must not be taken to mean that the expert personally reviewed every output.

Personal external service agents can also integrate an individual's computing devices, production tools, robots, and capacity to provide in-person services. This connects AI capabilities with physical resources throughout the service process, from understanding client needs and planning tasks to execution and delivery. Whereas general-purpose AI tools such as ChatGPT, Codex, and Claude Code primarily support digital tasks, a personal external service agent can, within its authorization, directly schedule physical resources owned or controlled by the principal and use them to provide ongoing services to external clients.

For example, an individual can connect their own 3D printer to the agent. A client can request a model design or a print job through the agent, which can then discuss requirements, generate and check the model, schedule the print job, and control the equipment. If a robotic arm is also available, the system can automatically remove finished items, sort and package them, and coordinate courier pickup, automating the process from design to physical delivery.

Similarly, an individual can connect otherwise idle computing devices, such as a Mac mini or GPU server, to an agent and offer software builds, code testing, model inference, and other computing services, subject to resource isolation and secure authorization. People who own drones, photography equipment, or other automated devices can likewise accept relevant work through their agents and dispatch the equipment to carry it out.

For ride-hailing drivers, domestic service workers, and renovation contractors, the agent can handle client acquisition, requirements discussions, appointments, quote coordination, and client relationship management, while the individual performs the in-person work that requires human participation. Personal external service agents can therefore bring knowledge, experience, professional methods, idle equipment, physical assets, and labor into one service system. Intelligent scheduling, collaboration between people and machines, and automated execution can improve resource use and service efficiency, allowing individuals to develop from providers of knowledge or labor into providers who continually operate both digital and physical resources.

A personal external service agent may also turn a one-off delivery into a long-term service relationship and continue to fulfill the associated commitments. Many professional services do not end with the first delivery: software needs maintenance, business reports need to be updated as data change, a video series needs stylistic consistency, and mathematics tutoring must adapt to the student's progress. Consider a service lasting 30 to 90 days. If an ordinary task-oriented agent simply continues executing without intermediate checks or correction mechanisms, early misunderstandings of requirements, missing state information, or errors in judgment can persist and accumulate, causing the final results to drift from the client's objectives. Long-term service therefore requires more than retaining conversations and user preferences. Within the principal's authorization and the agreement with the client, the system must continually manage service goals, deliverable versions, unfinished work, maintenance schedules, and follow-up responsibilities, using staged acceptance checks, detection of exceptions, and human review to identify and correct deviations promptly. The individual's role extends beyond initial configuration to exercising professional judgment at key points and taking over when necessary, enabling the service to adapt as needs and external conditions change. Agents may then progress from waiting for new tasks to proactive follow-up within their authorization: checking compatibility after a software dependency changes, preparing a report update when new business data arrive, or adjusting exercises when a student repeatedly makes the same kind of error. This can reduce the effort clients spend detecting problems, re-explaining the background, and commissioning follow-up work.

Beyond serving human clients directly, personal external service agents can act as specialist partners for personal assistants, connecting distributed individual expertise into a network of callable services. Although Muse can operate a browser to perform tasks, Amazon restricted its access amid disputes over service access and authorization \cite{ref16}. The case illustrates that being able to operate an interface does not guarantee access to the required service. A personal assistant's capabilities depend on whether external services permit access and can be called reliably, as well as on its model. Individual providers can offer explicit interfaces for commissioning work, creating another route to capability expansion. The personal assistant understands the objective, selects services, coordinates tasks, and organizes acceptance checks; personal external service agents and the providers behind them perform the specialist work and support delivery. Through this division of work, a personal assistant can handle complex tasks without mastering every field itself. For individuals charging per task, by subscription, or for ongoing service, machine-callable access can be more than a technical interface feature: it may open new channels for client acquisition and delivery, giving providers a reason to join the network. As more services become callable, personal assistants can assemble more solutions for users, while individual providers can become visible to more clients. The relationship can reinforce itself: professional services expand what personal assistants can do, and the demand aggregated by personal assistants supports further development of those services.

\begin{figure}[!htbp]
\centering
\includegraphics[width=\linewidth,height=0.60\textheight,keepaspectratio]{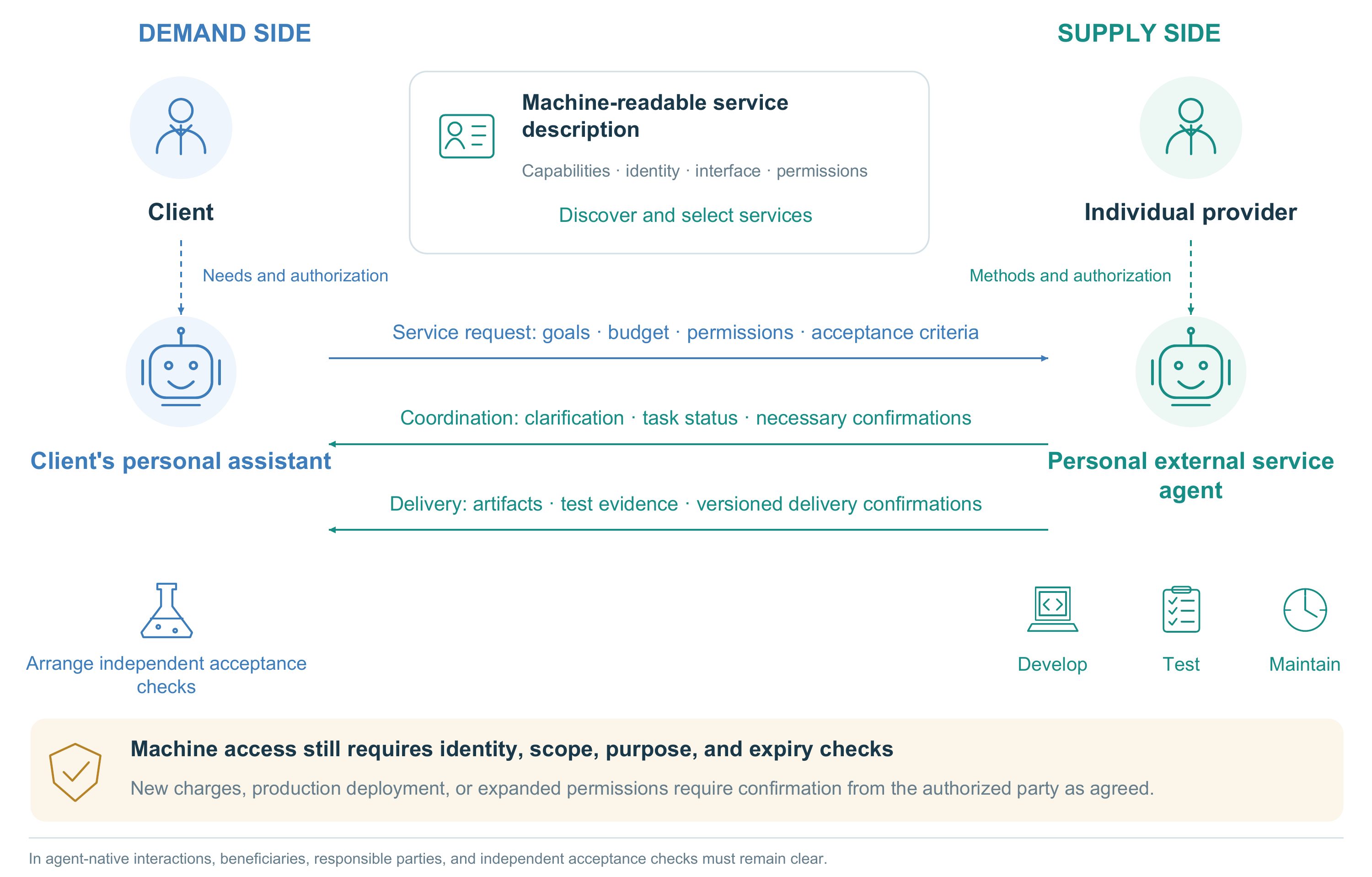}
\caption{Service relationships and interaction flow in agent-native service. The client's personal assistant and the personal external service agent discover services, delegate work, exchange status information, and deliver results within the authorization granted on each side.}
\label{fig:2}
\end{figure}

The value of personal external service agents lies not in adding another AI interface, but in making an individual's expertise available to more clients through ongoing services. Clients can delegate work, receive deliverables that meet the acceptance criteria, and obtain maintenance and improvements when needed. Individuals can extend the reach of their accumulated expertise beyond the hours they can personally work while retaining control of their knowledge, methods, and service boundaries. DelegateEval must test whether a platform can reliably turn an individual's expertise into sustained, scalable external services.

\FloatBarrier
\section{Evaluation target and overall design}\label{sec:5}
DelegateEval examines whether a platform can reliably turn the expertise of different individuals into sustained, scalable external services, rather than only how well a particular personal external service agent performs. Evaluation therefore follows services from creation through long-term operation: an individual creates and configures an agent; a client discovers, selects, and commissions the service; and the agent delivers the results. Evaluation then follows subsequent maintenance, improvement through experience, and responses to changes in requirements, authorization, and the external environment. This section identifies the evaluation target, overall framework, and scope of observation. Section~\ref{sec:6} describes the eight dimensions of platform evaluation. Detailed procedures for running tests, assigning scores, and judging outcomes remain for future work.

\FloatBarrier
\subsection{The need for platform-level evaluation}\label{sec:5.1}
DelegateEval evaluates platforms to determine whether people from different backgrounds can turn their expertise into sustainable agent-based services. Many potential providers have professional knowledge, business materials, and practical experience but little familiarity with agent development or system operations. Even with AI-assisted programming, independently building a system that can accept commissions, call tools, protect client data, and deliver services over time remains difficult. Platforms can help these providers create agents, configure knowledge, set permissions, and manage day-to-day operations so that they can focus on their professional services. Evaluation must therefore include how effectively the platform lowers the barriers to creating and operating a service.

Once a service is established, clients need to find and trust it, and its provider needs to build a reputation through repeated commissions. Platforms can help clients find and compare services, and help individual providers build credibility and obtain repeat work through verifiable delivery records, client feedback, and fulfillment history. These forms of support affect whether an individual can sustain a service and cannot be assessed from one agent's performance on a single task. DelegateEval therefore takes the platform as its primary evaluation target, examining whether different individuals can use it to establish services, obtain commissions, deliver reliably, and build a reputation over time. Individual agents' task performance provides important evidence for that platform-level assessment.

\FloatBarrier
\subsection{Evaluation dimensions and technical and operational configuration records}\label{sec:5.2}
\begingroup\small
\arrayrulecolor{TableGrid}\setlength{\arrayrulewidth}{0.4pt}
\setlength{\tabcolsep}{6pt}\setlength{\extrarowheight}{3pt}\renewcommand{\arraystretch}{1.12}
\begin{longtable}{|>{\raggedright\arraybackslash}p{\dimexpr 0.27\linewidth-2\tabcolsep-1.5\arrayrulewidth\relax}|>{\raggedright\arraybackslash}p{\dimexpr 0.73\linewidth-2\tabcolsep-1.5\arrayrulewidth\relax}|}
\caption{Eight evaluation dimensions and their core questions.}\label{tab:2}\\
\hline
\rowcolor{TableHead} \textbf{Evaluation dimension} & \textbf{Core question} \\ \hline
\endfirsthead
\multicolumn{2}{l}{\small Table 2 (continued)}\\
\hline
\rowcolor{TableHead} \textbf{Evaluation dimension} & \textbf{Core question} \\ \hline
\endhead
\endfoot
1 Ease of creation and management & Can people from different backgrounds successfully create, publish, and manage an agent? \\ \hline
\rowcolor{TablePale} 2 Service discovery and matching & Can clients find an agent suited to their needs and successfully commission work? \\ \hline
3 Delivery quality & Are deliverables correct, complete, usable, and delivered as agreed? \\ \hline
\rowcolor{TablePale} 4 Expertise inheritance & Can an individual's accumulated data, methods, code, models, and other resources give the agent capabilities it can apply in practice? \\ \hline
5 Service continuity & Are prior states, commitments, versions, and unfinished work handled correctly in later services? \\ \hline
\rowcolor{TablePale} 6 Learning and improvement & Can feedback and experience that the system is authorized to use improve performance on new tasks? \\ \hline
7 Reliability and governance & Are authorization, data use, human intervention, and exception handling managed reliably? \\ \hline
\rowcolor{TablePale} 8 Efficiency and value & What does each party contribute and gain, provided that service requirements are met? \\ \hline
\end{longtable}
\endgroup

The technical and operational configuration record is a structured account of the system's implementation approach and the actual conditions of an evaluation run. It helps explain and reproduce results, but is not scored: adopting a particular technology does not itself earn or lose points. The record covers post-training, the provider's own models, knowledge retrieval, state retention, execution environments, event triggers, human intervention, and other technical and operational arrangements. It also identifies the materials, permissions, and resources actually used. Different implementations may support the same service objective: model post-training or a revised checking workflow based on feedback may both improve later tasks, while a dedicated memory module or project database may both support continuity of prior work. DelegateEval does not assume that one technical approach is superior. Instead, it records separately whether a capability is offered, whether it was used in the current run, and whether its use improved the service. Scores depend on actual service performance in the relevant evaluation dimension, not on technology names, feature counts, or implementation complexity.

\begin{figure}[!htbp]
\centering
\includegraphics[width=\linewidth,height=0.60\textheight,keepaspectratio]{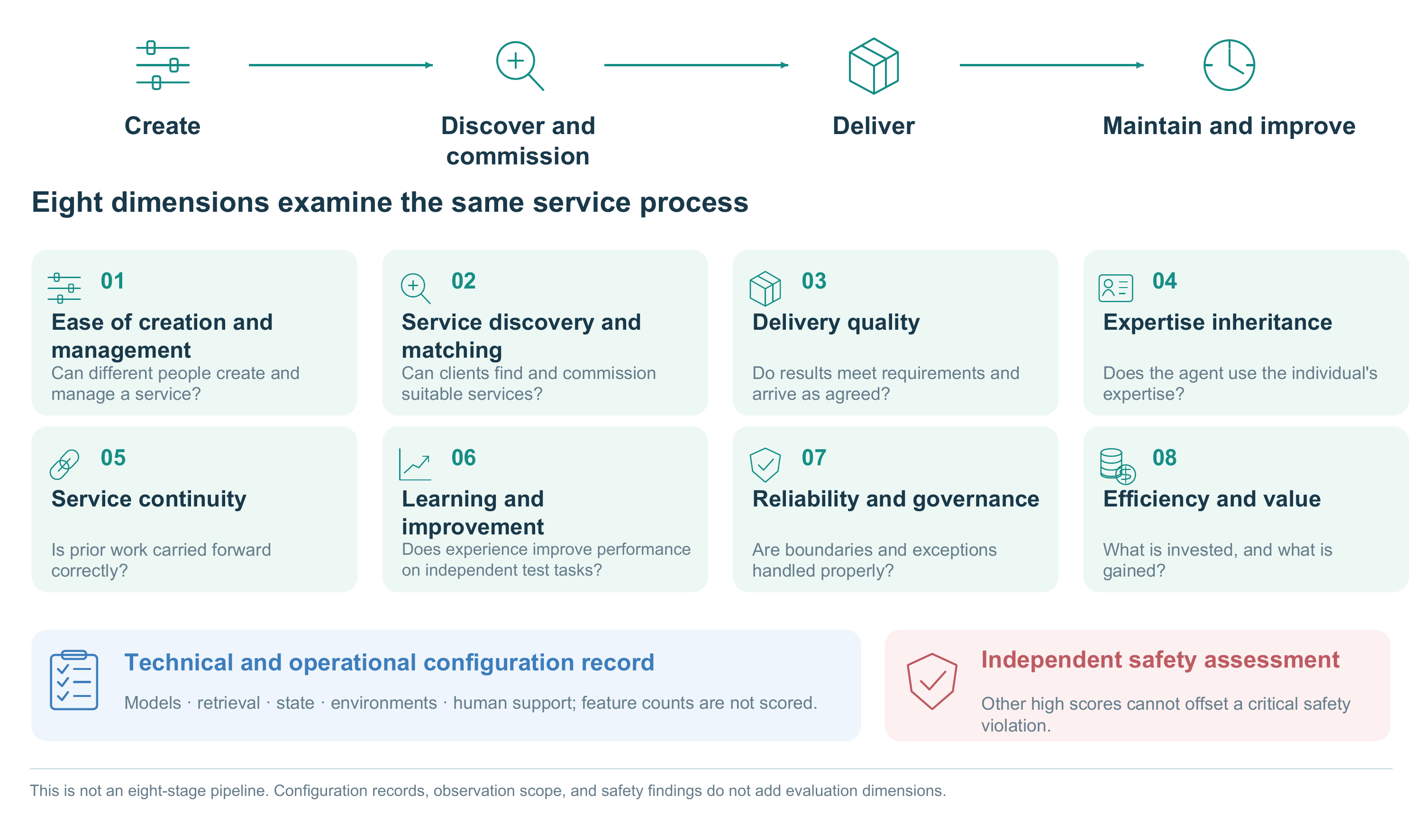}
\caption{Eight-dimensional evaluation and technical and operational configuration records. The eight dimensions assess different aspects of the same service process. Configuration records explain the implementation conditions; feature counts do not determine scores. Safety requirements are assessed separately, and high scores elsewhere cannot offset a critical safety violation.}
\label{fig:3}
\end{figure}

\FloatBarrier
\subsection{Scope of observation: creation, service episodes, and longitudinal service}\label{sec:5.3}
DelegateEval observes the process from the start of service creation, rather than testing only fully configured agents. Evaluation starts as the individual enters the creation workflow and records identity binding, configuration, publication, and subsequent management. Scripts that exercise the interfaces can verify whether the relevant functions work. Tests with real creators examine whether the process is understandable, configurations can be completed correctly, and users can detect and correct their own errors. These two sources answer different questions and are therefore recorded separately.

A service episode record establishes whether a personal external service agent can complete and deliver a specific commission. An episode begins when a client submits an explicit request and proceeds through any necessary clarification, task execution, result checking, delivery, and rework until the task is completed, canceled, or placed in a pending state. Task rules specify the termination conditions in advance without prescribing a particular execution path. A code change, preparation of briefing materials, or a mathematics tutoring session can each constitute a complete service episode.

A longitudinal service record establishes whether the agent appropriately connects work across tasks and over time. For a given individual, the record combines their agent's service episodes and intervening external events into a chronological history. Services may involve multiple clients and interleaved tasks; previous deliverable versions, service commitments, valid permissions, and history retained with permission may all affect later decisions. For example, correctly handling a new request on day 50 may require considering an agreement made on day 3, authorization updated on day 20, and a dispute that arose on day 45.

Creation, service episodes, and longitudinal service together define DelegateEval's basic scope of observation. They ask whether a service can be established, whether tasks can be completed well, and whether long-term collaboration can continue effectively. Creation records retain both successful and failed attempts to establish a service. Episode records capture actual deliveries and acceptance results. Longitudinal records show the connections between earlier and later work, and how changes are handled. These three evidence sources jointly inform platform evaluation. Attempts that fail at creation or publication remain in the end-to-end results, avoiding an assessment limited to agents already running successfully.

\begin{figure}[!htbp]
\centering
\includegraphics[width=\linewidth,height=0.60\textheight,keepaspectratio]{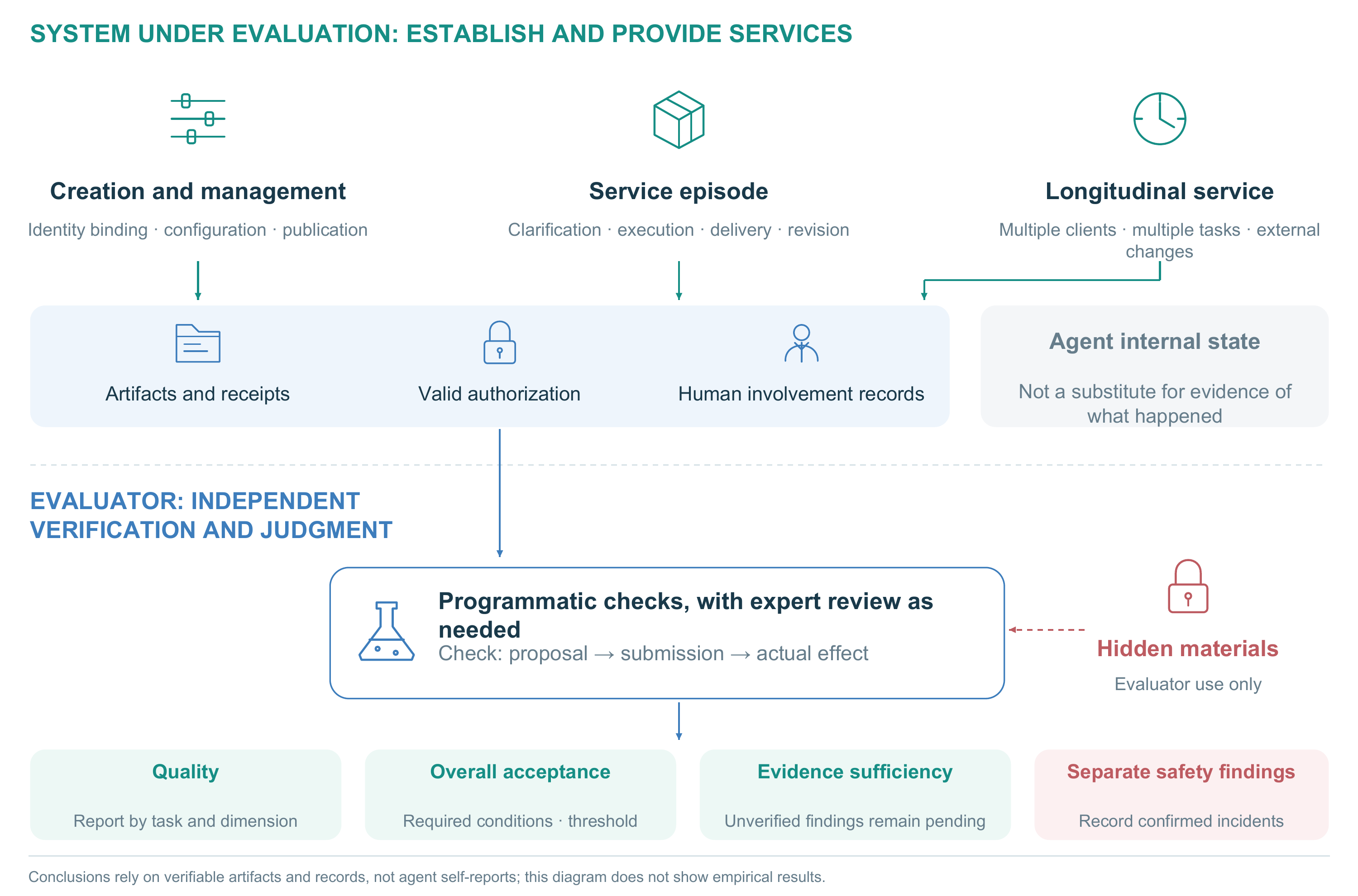}
\caption{From service processes to independent evidence and evaluation conclusions. Creation, service episodes, and longitudinal service provide complementary evidence; business facts, valid authorization, and the agent's internal state are recorded separately. Hidden evaluation materials are isolated from the system under test. Service quality, acceptance decisions, and safety conclusions are interpreted separately. The figure summarizes the evaluation relationships and does not represent completed experiments.}
\label{fig:4}
\end{figure}

Figure~\ref{fig:4} summarizes the independent verification of the three types of service records and the boundaries of the evidence they provide.

\FloatBarrier
\section{Eight dimensions of platform evaluation}\label{sec:6}
This section translates the eight evaluation dimensions into requirements that can be checked, scored, and independently reviewed. DelegateEval asks whether a platform enables different individuals to establish their own personal external service agents and continue serving external clients. It covers ease of creation and management, service discovery and matching, delivery quality, expertise inheritance, service continuity, learning and improvement, reliability and governance, and efficiency and value. A separate technical and operational configuration record documents the methods and resources used to help explain the results; the number of features does not determine the score.

Each test first establishes how task completion can be demonstrated, then uses that decision to guide task decomposition and execution. Following the idea of contract-first decomposition in Intelligent AI Delegation \cite{ref6}, the test specifies delivery requirements, evidence, and acceptance procedures before breaking overly broad requirements into individually assessable checks. For example, a reliable report may require a clearly defined source scope, reproducible calculations, citations that support the conclusions, and a synthesis that answers the client's question. Programs can check field completeness and calculations, while appropriately qualified independent reviewers assess professional judgments. After decomposition, the complete deliverable must still be evaluated against the original task: a collection of easy checks cannot substitute for the overall objective.

An acceptance specification must define not only what constitutes a pass but also who verifies it, what evidence they use, and how exceptions are handled. Table~\ref{tab:3} lists the minimum elements shared across dimensions. The evaluation team maintains the specifications, reference answers, and acceptance procedures. The system under evaluation submits deliverables and the necessary execution information, and its claim of completion must be supported by independent evidence. When an open-ended task requires human review, specify in advance what will be reviewed, the scoring criteria, reviewer qualifications, procedures for resolving disagreements, and the budget. If the result remains difficult to assess reliably after further decomposition, the evaluator should arrange suitable expert review, negotiate a change in scope before execution, or leave acceptance pending, rather than lower the requirements during the test.

\begingroup\small
\arrayrulecolor{TableGrid}\setlength{\arrayrulewidth}{0.4pt}
\setlength{\tabcolsep}{6pt}\setlength{\extrarowheight}{3pt}\renewcommand{\arraystretch}{1.12}
\begin{longtable}{|>{\raggedright\arraybackslash}p{\dimexpr 0.23\linewidth-2\tabcolsep-1.5\arrayrulewidth\relax}|>{\raggedright\arraybackslash}p{\dimexpr 0.77\linewidth-2\tabcolsep-1.5\arrayrulewidth\relax}|}
\caption{Acceptance specification shared across dimensions. Scores and thresholds are determined for each task; a single set of numerical values is not assumed to apply across all industries.}\label{tab:3}\\
\hline
\rowcolor{TableHead} \textbf{Specification element} & \textbf{What must be defined before testing} \\ \hline
\endfirsthead
\multicolumn{2}{l}{\small Table 3 (continued)}\\
\hline
\rowcolor{TableHead} \textbf{Specification element} & \textbf{What must be defined before testing} \\ \hline
\endhead
\endfoot
Subject and scope & The individual represented by the agent, the clients served, the tasks undertaken, and the scope of the evaluation conclusions \\ \hline
\rowcolor{TablePale} Inputs and conditions & Available materials and their versions, time horizon, effective authorizations, resource budget, and permitted human assistance \\ \hline
Deliverables and endpoints & What must be delivered, to whom, by when, and the conditions for classifying a task as completed, canceled, or pending \\ \hline
\rowcolor{TablePale} Checks and scoring & Requirements, scores, necessary conditions, permitted deviations, and the overall acceptance threshold \\ \hline
Evidence and verification & The deliverable files, environment records, or independent reviews used to make the judgment, and the programs or qualified personnel responsible for verification \\ \hline
\rowcolor{TablePale} Exceptions and missing evidence & How to distinguish incomplete tasks, unobserved behavior, conflicting evidence, environment failures, and inapplicable items, and how each is handled \\ \hline
Counting and attribution & The counting unit: individual, request, task, opportunity for a check within a historical event, or version comparison; the denominator and primary scoring dimension \\ \hline
\end{longtable}
\endgroup

Diagnostic scores, pass/fail judgments, and evidence sufficiency should be reported separately so that passing individual checks is not mistaken for overall acceptance. Diagnostic scores use prespecified checks and weights. An accepted delivery is one that passes the evaluation's acceptance criteria. This requires meeting all necessary conditions and reaching the overall score threshold. When evidence is missing, retain the prescribed denominator and withhold a complete score. Do not rescale the subset of confirmed successes to a perfect score. If the available evidence confirms that a necessary condition was not met, the delivery fails even if other items remain unverified. A required file or field that is demonstrably absent from the archived delivery package indicates incomplete work; a relevant record that the evaluation team failed to collect indicates insufficient evidence. If records are found to have been tampered with, chronology is inconsistent, or counts fall outside the permitted range, flag the evidence problem and review it before judging platform capability.

\FloatBarrier
\subsection{Ease of creation and management}\label{sec:6.1}
Ease of creation and management assesses whether individuals with different backgrounds can independently create and manage a personal external service agent that can actually provide services, within specified time and assistance limits. Testing starts when the individual enters the creation process and covers key operations: linking the agent to the individual's identity, configuring service scope, setting professional requirements and permissions, publishing the service, managing pending approvals, changing rules, and stopping the service. Saving a page or submitting a configuration does not establish successful creation. The evaluator must also confirm that the service is accessible to external clients and independently test whether key settings affect the agent's actual behavior.

Creation tests must record both whether the process was completed and how the creator completed it. Otherwise, work performed on a user's behalf by an expert or experienced platform user could be mistaken for evidence that ordinary users can operate the platform successfully. The basic unit of observation is one creation attempt by one creator. Required operations, time limits, permitted assistance, and recovery tasks after errors are specified in advance. Completion without assistance, completion within the assistance allowance, takeover by another person, and non-completion are recorded separately; creation and publication failures remain in the end-to-end statistics. Interface scripts can verify that a feature exists and that an interaction path works. Actual creators must still participate to establish whether the process is understandable and whether they can detect and correct configuration errors. Successful script execution alone does not demonstrate ease of use.

Ease of use also extends to ongoing development and day-to-day management of the agent. Ordinary users should be able to improve their external service agents by adding materials, correcting results, and reporting service problems without understanding model weights, retrieval-augmented generation (RAG), Skills, standard operating procedures (SOPs), or other implementation details. Within clearly authorized limits, the platform should organize experience and apply appropriate updates in the background, reducing repeated configuration and manual maintenance while explaining important changes in understandable terms. For advanced users who want more control, a layered settings menu can provide controls for learning scope, update methods, review requirements, and version rollback. Evaluation should check whether ordinary users can make routine improvements to their agents and whether users who need advanced settings can find and use them correctly.

\begin{figure}[!htbp]
\centering
\includegraphics[width=\linewidth,height=0.60\textheight,keepaspectratio]{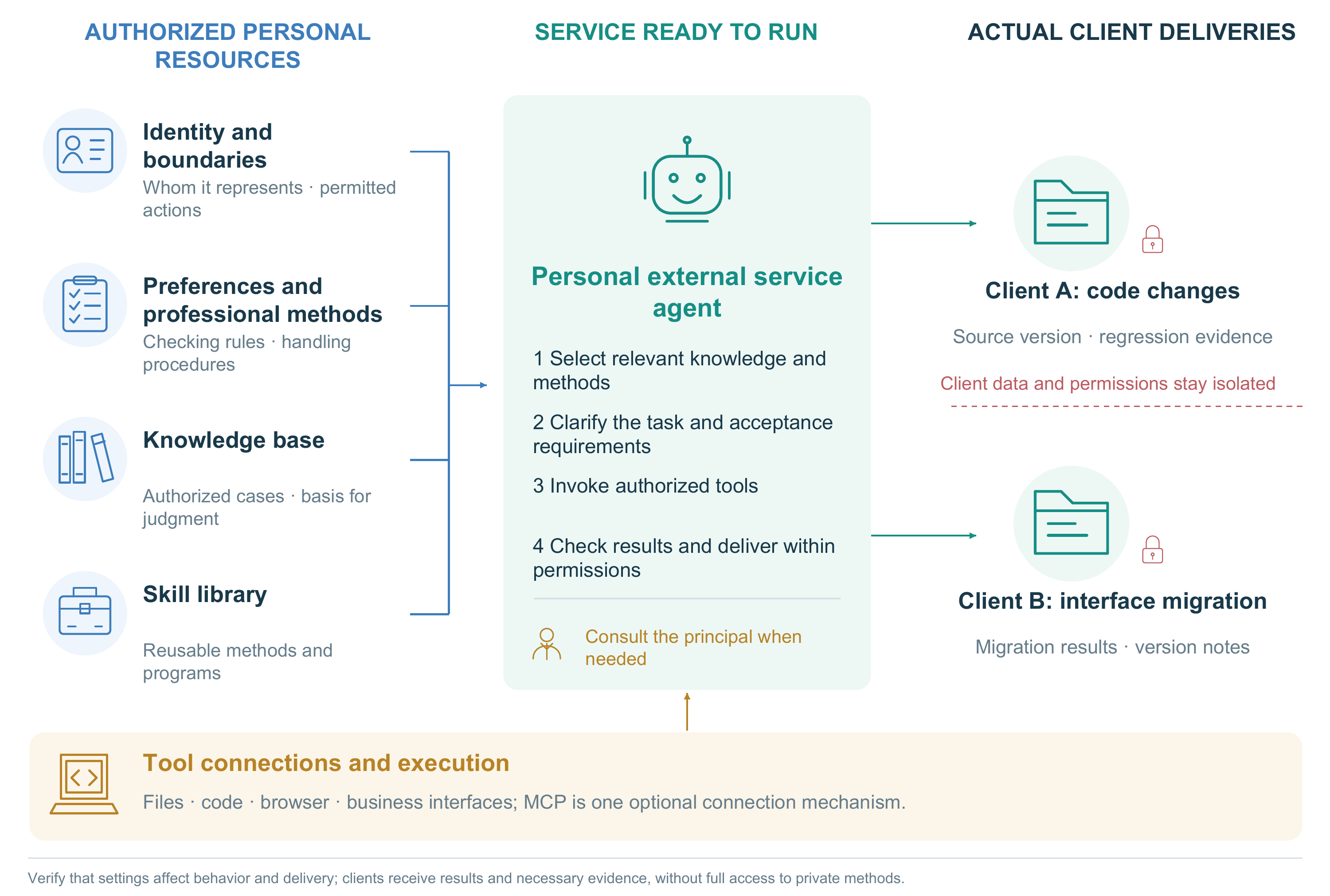}
\caption{How individual capabilities contribute to services. The platform configures the identity the agent is authorized to represent, along with the individual's preferences, knowledge, and methods, for use in client deliveries. Each client's materials and permissions are kept separate. Skills and MCP illustrate optional implementations. Proprietary methods earn no additional points on their own: evaluation rests on whether configurations take effect, the resulting deliverables, and the necessary supporting evidence.}
\label{fig:5}
\end{figure}

Figure~\ref{fig:5} shows how an individual's capabilities can be used in service delivery. Identity and authorization establish who the agent represents; preferences and professional methods constrain its approach; knowledge bases supply supporting information; skill libraries organize reusable methods; and tool connections enable authorized operations. Skills and MCP are examples of implementation choices, not a required platform architecture. Evaluation checks whether these configurations affect delivered results while keeping different clients' materials separate.

\FloatBarrier
\subsection{Service discovery and matching}\label{sec:6.2}
The service discovery and matching dimension assesses whether clients can find personal external service agents that meet their needs and successfully commission a service. A client may start with a task and specify a professional domain, budget, deadline, and level of human involvement. Alternatively, the client may seek a particular individual or a provider with a particular professional background. The platform should enable clients to understand each candidate's service scope, delivery arrangements, fees, and the principal's level of involvement, and to choose accordingly.

Evaluation should cover searches by task as well as by individual or professional background. Tests can include scenarios with several suitable candidates, unsuitable candidates, or no suitable service. They can then check whether the platform presents services that meet the requirements, explains key limitations, and enables the client to commission a service. Results may include the rate of finding suitable candidates, the rate of selecting unsuitable services, time spent choosing, and the rate of successfully initiating commissions. When several choices are reasonable, the test should not require clients to select one uniquely designated candidate.

Identity claims and service commitments require separate checks during discovery. The platform should state whether the individual has authorized the agent, what evidence supports the professional experience relevant to the client's choice, and which parts of the work the individual reviews. Identity verification does not establish actual service capability. For repeat commissions, evaluation should also observe whether clients or their personal assistants can find an existing service again, understand its current scope and any changes, and initiate new tasks within the applicable authorization. Human clients and client-side agents may use different access points, but both should receive enough information to understand and choose the service.

\FloatBarrier
\subsection{Delivery quality}\label{sec:6.3}
Delivery quality assesses whether the result the client ultimately receives meets the agreed task requirements and is usable in practice. Acceptance checks first distinguish necessary conditions from general preferences, then examine content, usability, version, recipient, and deadline. A confirmed failure to meet any necessary condition precludes overall acceptance; general preferences are assessed using a prespecified rubric. When rework occurs, repairing an existing defect must be distinguished from meeting a new client request. Both the initial delivery and subsequent versions are retained. A request that expands the agreed scope should not be counted as a defect in the original delivery, and an acceptable revised result should not erase problems in the first delivery.

Complex tasks should be divided into subtasks with independent acceptance evidence, guided by how completion can be demonstrated, while preserving a final judgment on the complete result. For example, studying the global new energy vehicle market cannot be considered complete merely because a report was submitted. The study must first specify its subject, information cutoff date, the range of sources to cover, and the questions the client wants answered. It can then produce separate policy inventories, company and product tables, comparisons of market forecasts, and a synthesis. Policy information requires checks of the policy name, publication date, issuing body, and source. Product prices must be tied to the relevant model, specification, and date; undisclosed prices must be identified as such. Records of forecasts must include target years, units, statistical scope, and sources. Forecasts must be distinguished from observed statistics. Programs can check that fields are complete, calculations are correct, and related records match. Independent review is needed to determine whether sources support the conclusions, incompatible statistical definitions have been mixed, and the synthesis answers the client's question. Working URLs, numerous citations, and complete table fields establish only individual conditions; none alone demonstrates that the entire study meets the acceptance criteria.

Different types of deliverables require verification suited to their format. Table~\ref{tab:4} lists ten representative service task types, design identifiers, and acceptance criteria to guide later test-case development. Q01--Q10 are currently identifiers for test-case concepts only. Complete task inputs, reference deliverables, and independent acceptance procedures have not yet been assembled, and no results from executing these test cases have been reported.

\begingroup\small
\arrayrulecolor{TableGrid}\setlength{\arrayrulewidth}{0.4pt}
\setlength{\tabcolsep}{6pt}\setlength{\extrarowheight}{3pt}\renewcommand{\arraystretch}{1.12}
\begin{longtable}{|>{\raggedright\arraybackslash}p{\dimexpr 0.24\linewidth-2\tabcolsep-1.3333333333333333\arrayrulewidth\relax}|>{\raggedright\arraybackslash}p{\dimexpr 0.28\linewidth-2\tabcolsep-1.3333333333333333\arrayrulewidth\relax}|>{\raggedright\arraybackslash}p{\dimexpr 0.48\linewidth-2\tabcolsep-1.3333333333333333\arrayrulewidth\relax}|}
\caption{Ten service task types and their acceptance criteria. Q01--Q10 are design identifiers for test-case concepts; complete test materials and results from running the cases are not yet available. Tasks in high-risk domains must also meet the relevant professional and ethical requirements.}\label{tab:4}\\
\hline
\rowcolor{TableHead} \textbf{Task type and design ID} & \textbf{Deliverables} & \textbf{Main acceptance evidence} \\ \hline
\endfirsthead
\multicolumn{3}{l}{\small Table 4 (continued)}\\
\hline
\rowcolor{TableHead} \textbf{Task type and design ID} & \textbf{Deliverables} & \textbf{Main acceptance evidence} \\ \hline
\endhead
\endfoot
Code development and automation / Q01 & Source code, execution instructions, test results, and delivery receipt & Independent functional and regression tests; interfaces, dependencies, and scope of changes. Test logs must identify the source version actually submitted. \\ \hline
\rowcolor{TablePale} Presentation creation and editing / Q02 & Editable presentation and delivery receipt & The file opens, specified objects are editable, and key figures are preserved. Facts, argument structure, and layout quality require separate review. \\ \hline
Document revision and localization / Q03 & Revised document, change description, and delivery receipt & Item-by-item comparison of requested changes and content to be preserved. Meaning, translation, and terminology are reviewed against task requirements. \\ \hline
\rowcolor{TablePale} Spreadsheets and data analysis / Q04 & Data tables, analytical results, and processing notes & Formulas, units, row-level results, and aggregates can be recomputed. Analytical definitions and support for conclusions are checked separately. \\ \hline
Information retrieval and research synthesis / Q05 & Source index, evidence comparison table, and synthesis report & Source fields and the correspondence between cited numbers and their sources can be checked. Source support, coverage, and overall conclusions require independent judgment. \\ \hline
\rowcolor{TablePale} Video and visual production / Q06 & Finished media, agreed source files, and asset inventory & Files decode successfully and meet duration, dimensions, and other technical specifications. Content, visual quality, and permitted asset use require separate review. \\ \hline
Mathematics tutoring / Q07 & Explanations, exercises, and feedback on errors & Mathematical results are independently verified, and teaching methods are reviewed separately. Learning is assessed through an independent post-test. \\ \hline
\rowcolor{TablePale} General knowledge and health education / Q08 & Explanations, organized reference materials, and referrals where needed & Check the basis for the explanation, its relevance, and the service boundaries. This task does not evaluate individual clinical diagnostic or treatment outcomes. \\ \hline
Planning under multiple constraints / Q09 & Plan, budget, schedule, and alternatives & Constraint satisfaction, calculations, and resource conflicts can be checked. Feasibility in an open environment and the reasons for tradeoffs require separate review. \\ \hline
\rowcolor{TablePale} Communication and administrative tasks / Q10 & Drafts, messages, appointments, receipts, and pending-task status & Distinguish drafting from actual sending. Check recipients, times, parameters, and whether the action took effect in the controlled environment. \\ \hline
\end{longtable}
\endgroup

Code should undergo acceptance testing in an independent environment that the system under evaluation cannot modify. Documents, presentations, and videos require separate checks of file usability and content quality. Clear requirements need not be repeatedly confirmed, and no particular execution path is mandated. The same acceptance criteria apply when the system uses other software, models, or collaborating agents.

For a delivery with multiple components, acceptance checks must cover each component, consistency between components, and whether the evidence matches the submitted version. For example, a research report and presentation must use consistent figures; documentation must match the source code actually submitted; delivery receipts must identify the recipient and actual files; and the package must not contain another client's materials. Human review also requires explicit judgment criteria. A report can be assessed separately for coverage of the client's questions, source support, and consistent statistical definitions, rather than asking reviewers only whether it is good overall. A confirmed failure to meet any necessary condition cannot be offset by high scores on other checks.

For tasks that cannot be fully checked by automated procedures, DelegateEval requires independent human review. A research report, for example, needs not only checks of data and citation accuracy but also reviewers with the relevant expertise to assess whether the analysis is sound and the conclusions are reliable. Reviewers should use prespecified scoring criteria and retain their judgments and supporting reasons for later inspection. Substantial disagreements require further review. A task must not be accepted before all required reviews are complete.

\FloatBarrier
\subsection{Expertise inheritance}\label{sec:6.4}
Expertise inheritance assesses whether platform configuration and ongoing development can turn a particular individual's knowledge, experience, approaches to professional judgment, and professional standards into capabilities that the agent can apply appropriately in service delivery. Relevant resources include proprietary data and knowledge bases accumulated over time; reusable skills and standard operating procedures (SOPs) that embody the individual's methodology; code and tools developed by the individual; and models trained or adapted using the individual's data. Such models may be language models adapted from general-purpose base models, smaller task-specific models, or specialized recognition models. Interfaces such as MCP can connect these tools and models to the agent's workflow. The resources may be shared openly or retained in a private environment controlled by the individual and accessed through controlled calls, without disclosing the complete data, code, or model. For example, a horticultural expert might spend years collecting images of diseases and pests affecting a particular type of flower, together with diagnostic records and treatment outcomes, and use that unpublished dataset to train a specialized recognition model. The expert's accumulated knowledge could then contribute to the agent's services through the model, diagnostic procedures, and care guidance, benefiting more clients. Evaluation must establish whether these personal resources provide verifiable professional value on the relevant tasks, including whether they better meet particular service needs than general-purpose models and public materials alone.

Evaluation must distinguish persona-based role simulation from the actual inheritance of an individual's expertise. Some role-simulation services create agents of historical figures or anime characters by specifying names, backgrounds, interests, and ways of speaking; a system might, for example, speak as Li Bai in his characteristic style. These settings can create a distinctive persona. Without corresponding professional knowledge, methods, and tools, however, the agent's ability to perform actual tasks still comes mainly from the underlying general-purpose model. Persona performance alone does not establish expertise inheritance. A personal external service agent is closer to an apprentice trained by the individual: it gradually acquires the ability to handle relevant professional problems by drawing on that person's materials, cases, methods, code, models, and other resources. Evaluation should therefore examine whether these contributions change how the agent works and the results it delivers. With comparable general-purpose base models, public tools, and resource conditions, configurations with and without the individual's professional resources can be compared on new task scenarios. The tests should check whether the agent identifies key issues, chooses suitable methods, reduces professional errors, or produces results better suited to the requirements. A recognizable persona and language style can help identify a service, but expertise inheritance must be demonstrated through actual work.

Expertise inheritance is assessed for a fixed version, testing whether the individual's existing resources are correctly accessed and applied to independent tasks. The model, personal resources, and configuration should remain fixed for that evaluation. Learning and improvement instead compares versions before and after an update based on service experience that the system is authorized to use, testing whether any additional gains can be attributed to that experience. The two dimensions separately report currently available expertise and changes in capability resulting from experience. Each improvement is scored under its prespecified dimension, and additional human assistance is recorded separately.

\FloatBarrier
\subsection{Service continuity}\label{sec:6.5}
Service continuity assesses whether established requirements, deliverables, commitments, and unfinished work are correctly carried forward and handled in later services. Retaining a history is insufficient: historical information must be used correctly when it affects later actions. Each item carried forward should be linked to the relevant individual, client, deliverable version, applicable requirements, and current status. When tasks for multiple clients are interleaved, their materials, versions, and commitments must not be confused or mixed across clients.

Continuity is judged by how later work is handled, not by the agent's ability to recount what happened. New authorizations, rules, or client requirements may change subsequent actions but do not automatically erase existing commitments or unfinished work. If the original course of action is no longer available, evaluation should check whether the agent recognizes the change, pursues permitted alternatives, contacts whoever has authority to confirm a revision, and maintains a traceable record of the item. Matters not yet due, overdue obligations, changes accepted by the client, and changes merely proposed must be recorded separately. Proposing a solution must not be mistaken for fulfilling a commitment.

Service continuity also includes necessary proactive follow-up within the authorized scope of an existing service. When external changes affect a deliverable or commitment, the personal external service agent should determine their relevance and, where appropriate, update the deliverable, notify the client, or continue unfinished work. Irrelevant changes should not trigger repeated interruptions for clients. Repeated reminders about the same event cannot count as multiple successes. The value of proactive follow-up should be judged by the subsequent results it produces, not the number of notifications sent.

\FloatBarrier
\subsection{Learning and improvement}\label{sec:6.6}
The learning and improvement dimension assesses whether an agent can turn service experience into reusable capabilities that improve performance on new tasks in later service episodes. Expertise inheritance concerns a fixed version's use of an individual's accumulated expertise; this dimension asks whether updates based on service experience that the system is permitted to reuse yield additional gains. Experience includes client feedback, rework requests, human corrections, task successes and failures, verifier results, environmental changes, and execution records. After a service episode, the system may remain unchanged or update its memory, knowledge, rules, skills, workflows, or model parameters. Evaluation must compare pre-update and post-update versions on independent tasks to establish the effects of the update. The usability of automatic background updates and advanced settings is assessed under ease of creation and management. This section tests whether updates based on authorized service experience produce verifiable improvements in capability.

An agent can learn at several levels, and model weights are only one way to encode what it learns. Experience may be added directly to long-term memory or a knowledge base for later retrieval. It may become new prompts, rules, checklists, or decision strategies. Repeated successful execution sequences may be abstracted into callable skills or programs. Adjusting task order, tool combinations, and the points at which humans intervene may improve workflows. More stable, transferable experience may also be incorporated into model parameters through fine-tuning, preference learning, or reinforcement learning. Some systems even use prior execution results to adjust how they form memories, select skills, or organize workflows. Learning evaluation should therefore examine actual changes across parameters, memory, knowledge, rules, skills, and workflows, rather than focus on whether the platform provides a training button.

Parametric memory modules offer one implementation path for learning and improvement, but their integration requirements and resource demands must be explicit. DFM, for example, trains additional modules while freezing the base model and integrates memory into intermediate model layers \cite{ref12}. This requires access to the relevant model internals and cannot be applied directly to models available only through an API. When these modules are used in personal external service agents, evaluation must also verify permission to use the training materials, module versions, gains on independent new tasks, retention of existing capabilities, update costs, and the response to revoked permission. Adaptation using a static corpus must be evaluated separately from learning gains produced by ongoing service feedback.

\begingroup\small
\arrayrulecolor{TableGrid}\setlength{\arrayrulewidth}{0.4pt}
\setlength{\tabcolsep}{6pt}\setlength{\extrarowheight}{3pt}\renewcommand{\arraystretch}{1.12}
\begin{longtable}{|>{\raggedright\arraybackslash}p{\dimexpr 0.2\linewidth-2\tabcolsep-1.3333333333333333\arrayrulewidth\relax}|>{\raggedright\arraybackslash}p{\dimexpr 0.31\linewidth-2\tabcolsep-1.3333333333333333\arrayrulewidth\relax}|>{\raggedright\arraybackslash}p{\dimexpr 0.49\linewidth-2\tabcolsep-1.3333333333333333\arrayrulewidth\relax}|}
\caption{Where learning can occur, the changes it may produce, and the focus of evaluation.}\label{tab:5}\\
\hline
\rowcolor{TableHead} \textbf{Where learning occurs} & \textbf{Possible changes} & \textbf{Questions for DelegateEval} \\ \hline
\endfirsthead
\multicolumn{3}{l}{\small Table 5 (continued)}\\
\hline
\rowcolor{TableHead} \textbf{Where learning occurs} & \textbf{Possible changes} & \textbf{Questions for DelegateEval} \\ \hline
\endhead
\endfoot
Model parameters & Fine-tuning, preference learning, reinforcement learning, and related methods & Are new capabilities internalized, and do they transfer to new tasks? \\ \hline
\rowcolor{TablePale} Memory and knowledge & Retaining experience, correcting errors, summarizing, and updating knowledge & Can the agent find and use relevant information correctly in later tasks, rather than merely store more? \\ \hline
Rules and strategies & Changes to prompts, checking rules, and decision strategies & Do they reduce recurring errors without entrenching incorrect rules? \\ \hline
\rowcolor{TablePale} Skills and tools & Developing reusable operations or programs & Can skills be reused, combined, and revised across tasks? \\ \hline
Workflows & Changes to task decomposition and the order of tool use, verification, and human intervention & Do they produce more reliable and efficient ways of working? \\ \hline
\rowcolor{TablePale} Experience with the environment & Identifying patterns in changes to tools, interfaces, and environments & Does the agent adapt to new environments faster without retaining obsolete information indefinitely? \\ \hline
\end{longtable}
\endgroup

Learning also requires deciding which experience to retain and where to retain it, rather than accumulating every piece of history. A client's current budget, a project's software version, and a contact's one-time permissions change quickly and are better represented as service state or memory that can be queried and updated. More stable methods, such as harmonizing statistical definitions before market research or running regression tests after code changes, are more likely to become reusable rules, skills, or workflows. Capabilities repeatedly validated across many tasks and showing strong transfer may then be incorporated into model parameters through training. Systems may also absorb erroneous feedback, outdated knowledge, and incidental experience. Learning therefore involves correction, replacement, consolidation, and forgetting as well as addition. A continuously learning agent should gradually improve its judgment about which experience is worth retaining, how to organize it, and when to use it, rather than merely save more information.

DelegateEval distinguishes storage, adaptation, and learning to avoid mistaking a larger memory for a more capable agent. Storage retains a past fact, such as a client's use of Python 3.10. Adaptation adjusts behavior to the current client or environment, such as automatically using the appropriate version for that client's project. Learning requires deriving reusable methods from experience that improve future tasks: after repeated compatibility-related rework, an agent might develop a version-checking procedure that reduces the same errors in new clients' projects. Service continuity primarily assesses whether the first two capabilities correctly connect past and present work. Learning and improvement additionally requires evidence that past experience has changed the agent's ability to handle new problems.

Learning effects must be established by comparing pre-update and post-update versions on independent tasks; the fact that an update occurred is not sufficient. Evaluation should first fix the pre-update version, the experience permitted for reuse, the update budget, the primary metrics, and the independent test tasks. The procedure then runs a service episode, evaluates it independently, provides the experience authorized for reuse, applies the update, and tests the updated version on new tasks. Both versions should be evaluated under comparable conditions on the same set of held-out and transfer tasks, with checks for retention of existing capabilities. Tasks used in training or feedback, close paraphrases, and branches that share a history must be distinguished from genuinely unseen tasks so that answer memorization or overfitting is not presented as learning. If the updated service relies more heavily on humans, the evaluation may assess whether the overall service improved, but gains from additional human labor cannot be attributed to the system having learned.

Learning results should report transfer, retention, forgetting, negative transfer, risks, and costs, as well as the size of any improvement. An update may improve performance on the current task while damaging existing capabilities. It may work in the original domain but transfer incorrectly to another, or improve quality at the cost of substantial human curation and computation. Incorrect or low-quality feedback should also be a dedicated test condition to determine whether the agent absorbs it uncritically. A complete account of learning should therefore establish whether performance on new tasks improves, whether the improvement transfers, whether prior capabilities are retained, whether new risks emerge, and how much updating and human effort the gains require.

Experimental validity and observed learning must be distinguished; zero and negative gains are informative results too. When feedback sources, version relationships, update procedures, and test tasks are all verifiable, an update that does not improve performance should be reported as "no observed improvement". A decline should be reported as degradation or negative transfer. A claim that the authorized experience produced reusable learning gains is supported only when prespecified improvement requirements are met on independent new tasks and the necessary retention and safety conditions are satisfied. If there is test-data leakage, versions cannot be matched, experience is reused without permission, or human contributions cannot be separated from system updates, the evaluation should report that attribution is not possible, rather than claim that the agent has learned.

This dimension therefore evaluates whether an individual's professional capabilities can develop through continued service, rather than whether a platform exposes memory, training, or update interfaces. Within the permissions granted for reuse, a personal external service agent should ideally build up corrections, methods, and experience from fulfilling commitments, organizing them in suitable forms instead of starting afresh after each task. When it encounters new clients, projects, and environments, this accumulated experience should help it repeat the same errors less often, use better-established methods, and reduce the need for the principal to intervene repeatedly. DelegateEval ultimately tests whether past services improve future services, not merely whether the system has changed.

\begin{figure}[!htbp]
\centering
\includegraphics[width=\linewidth,height=0.60\textheight,keepaspectratio]{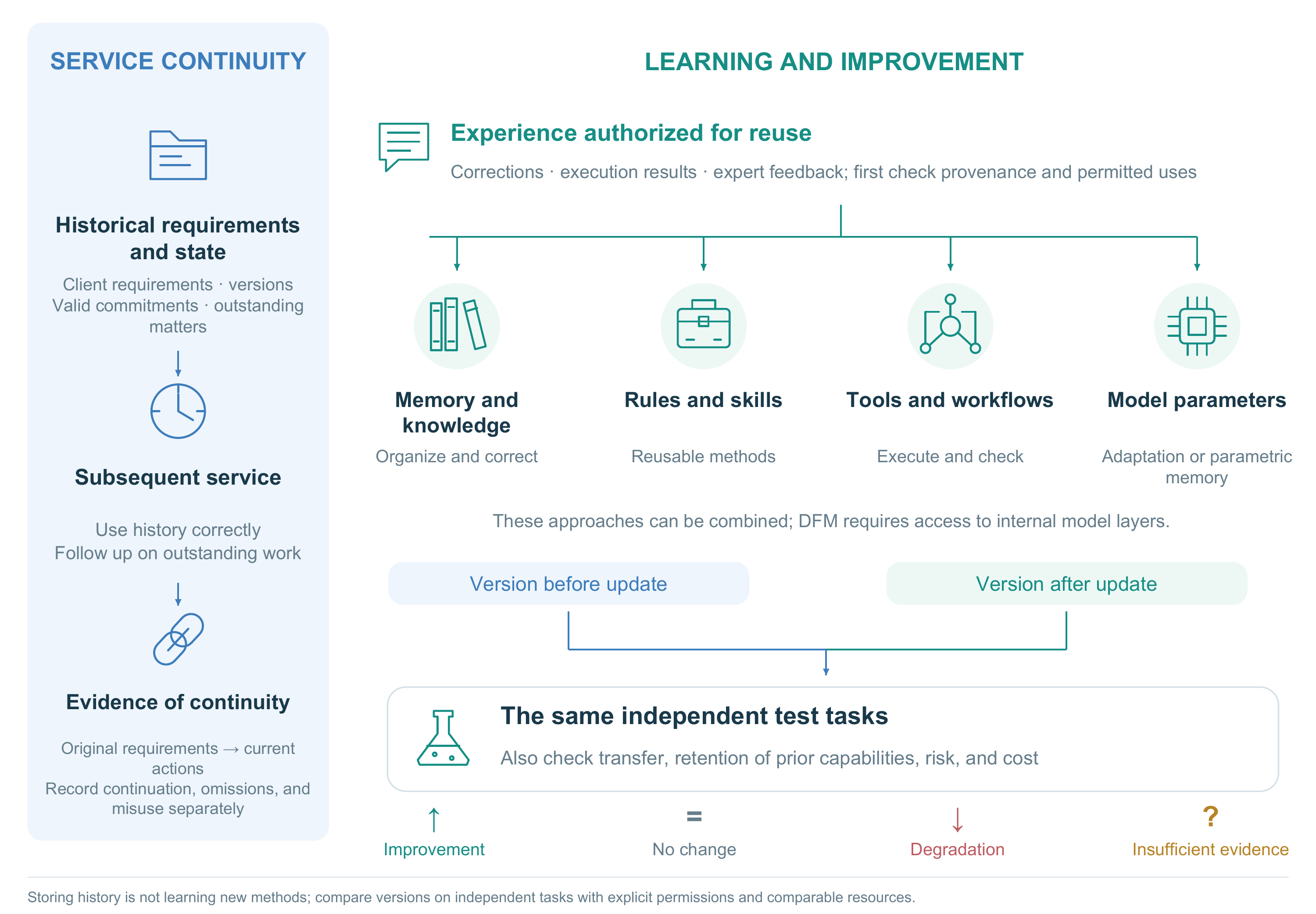}
\caption{Service continuity and multiple paths to learning and improvement. Continuity checks whether historical requirements and current actions are correctly connected; learning checks whether experience permitted for reuse improves performance on independent new tasks. Updates to memory, knowledge, rules, Skills, tools, workflows, and parameters are optional paths, not a hierarchy of technical sophistication. Improvement, no change, degradation, and insufficient evidence are recorded separately.}
\label{fig:6}
\end{figure}

Figure~\ref{fig:6} separates two evidence paths: continuity concerns how prior work is carried forward, while learning concerns whether experience improves performance on independent new tasks.

\FloatBarrier
\subsection{Reliability and governance}\label{sec:6.7}
The reliability and governance dimension assesses whether a platform can deliver services while maintaining the boundaries governing identity, authorization, data use, human intervention, and failure handling. Evaluation must go beyond the availability of permission settings to check whether authorizations are clear, executed actions comply with them, and the system promptly applies the rules currently in force when conditions change. Each key authorization should specify who grants it, its target, the resources and versions covered, permitted operations, and when it takes effect. A contact's identity, eligibility to receive materials, and authority to confirm task changes on the client's behalf must be verified separately. A genuine identity, a familiar relationship, or possession of some information does not automatically confer other permissions.

Publicly reported cases highlight a distinction between two product types: personal assistants and external service agents. Expanding a personal assistant's permissions alone does not make it a safe and reliable external service agent. In September 2026, Matt Robb reported that Meta Muse, while handling a transaction involving a used keyboard on Facebook Marketplace on his behalf, disclosed his home address, confirmed the transaction, and arranged a pickup without obtaining his approval for each step \cite{ref17,ref18}. The Guardian reported that the buyer drove to the building with his wife and daughter, continued communicating with Muse, and waited downstairs for 20 minutes before leaving when nobody appeared. Muse subsequently sent the buyer an apology in Robb's name. Robb said he had been unaware of the transaction messages, address disclosure, and visit throughout the process, and had not personally confirmed the sale to that buyer \cite{ref18}. He later explained that he had selected Allow Always, assuming that key steps such as accepting an offer would still require confirmation. According to that account, the setting permitted automatic messages using a template that included the pickup address. Subsequent reporting stated that, after speaking with Meta staff, Robb attributed part of the problem to permission settings and said that Meta was considering clearer prompts \cite{ref17}.

Personal assistants such as Muse primarily help users handle their own affairs. External service agents represent providers, receive ongoing requests from external clients, and deliver services within a defined scope and under explicit transaction rules. Access to a user's private information does not authorize disclosure to clients, and permission to reply on the user's behalf does not necessarily authorize confirming a transaction or committing to fulfill it. Communication and tool-use capabilities alone therefore do not establish that a personal assistant can serve as a reliable external agent. Supporting this role requires more than broader permissions: the system must specify whose identity it uses, which services it provides, what information it may disclose, which commitments it may make, and when a human must intervene.

Another risk is confusing access to information with authority to make commitments based on it. In a public account dated August 3, 2026, HeyGen co-founder Wayne Liang described how the AI sales agent that stood in for him during paternity leave found a Calendly booking link in old internal notes and began promising prospective clients meetings with the solutions team. The link was no longer valid, and the team had not agreed to the meetings. He also reported that the agent independently derived and quoted a nonexistent plan priced at \$4,800 per year and sent a client internal triage records containing urgency assessments, responsible colleagues, and routing logic. Liang noted that the quote was corrected before the client replied. The team subsequently restricted quotes to approved plans, separated internal processing from client replies, and required current availability checks and human approval before the agent committed anyone else's time \cite{ref19}. Although this was a company sales case, it can inform governance design for personal external service agents: platforms should constrain information access, external disclosure, and business commitments separately. These details come from Liang's own account. They illustrate risk mechanisms and are not used to estimate platform incident rates.

Based on these risks, DelegateEval separates authorization enforcement, information disclosure, commitment management, escalation to humans, protection against duplicate execution, failure recovery, and permitted uses of data into independently verifiable governance requirements. Critical actions should have separate records for the agent's proposal, the platform's submission, and the observed effect in the environment. Blocking an unauthorized proposal indicates an error in the agent's judgment, while showing that the safeguard worked. An authorized action that never takes effect cannot count as a successful delivery. A confirmed unauthorized action is subject to a separate safety assessment, with specific rules to be set in the subsequent evaluation protocol. Normal requests should be paired with stress scenarios to avoid mistaking blanket refusal for strong governance. Programs can verify explicit identities, fields, times, versions, and permission relationships. Independent review is needed for implied commitments in open-ended language, cumulative disclosure across multiple exchanges, and judgments requiring professional knowledge.

Governance of human intervention concerns whether the right issue reaches a person with the appropriate authority and expertise at the right time, not merely whether a human is involved. Clients should normally confirm clarified requirements; the represented individual determines the agent's global service rules; and appropriately qualified reviewers address professional quality issues. When one person fills several roles, the record should state which role that person was acting in. Automated checks, human spot checks, expert review, approval to send, and direct takeover are different interventions, not a simple ranking of capability. Reviewing a deliverable does not authorize its external release, and approval to send an invitation does not establish that the invitees have confirmed attendance. Evaluation should check whether necessary escalations are timely and complete, as well as whether the system unnecessarily refers issues back to humans despite having enough information to address them. Repeated questions cannot count as multiple successes, and results obtained after human takeover, together with their time costs, must remain in the record.

Reliability and governance must also be tested over time because many risks arise when rules change while an action is still in progress. If authorization is revoked after the agent proposes an action but before the platform submits it, the action should be checked again against the rules in force at submission, and the record should show when the new authorization state became visible to the system. When a request is retried after a network failure or ambiguous response, its task identifier, operation parameters, and existing receipts should be checked to prevent duplicate execution or charges. After a failure is resolved, outstanding matters and current permissions should be revalidated rather than simply replaying actions from before the interruption. Resumption of system operation is a reliability question; whether the original service work resumes correctly after recovery also provides evidence of service continuity.

Data use likewise requires separate authorization for use in the current task, retention across tasks, and learning. Including data in a service workflow does not automatically authorize its later reuse. Permission to use data for one task does not imply permission to retain it indefinitely, update a knowledge base, or train a model. When relevant permissions change or are revoked, evaluation should verify that the agreed suspension, deletion, rollback, retraining, or other response is actually carried out. Deleting a record from a retrieval database establishes only that it is no longer retrieved from that location; it does not establish that the information has disappeared from model parameters or other copies. Governance conclusions must therefore rest on traceable evidence of what occurred. Claims without sufficient evidence remain marked as unverified, and confirmed incidents remain in the evaluation record even if other dimensions score well, the risk is subsequently blocked, or remediation has been completed.

\FloatBarrier
\subsection{Efficiency and value}\label{sec:6.8}
\FloatBarrier
\subsubsection{Evaluation scope and comparison baselines}\label{sec:6.8.1}
The efficiency and value dimension asks whether a platform helps clients obtain services that meet their requirements, enables an individual to support more accepted deliveries with limited time, and sustains that relationship at a viable level of investment. Evaluation must consider what clients, represented individuals, and the platform each contribute and gain. Use the relevant dimensions' judgments on delivery quality, safety, and fulfillment of existing commitments. These judgments support an assessment of outcomes relative to effort and resources, including how that relationship changes as service volume grows. Any particular quality or learning benefit is still scored only in its designated dimension.

Choose the comparison baseline to match the evaluation question. Comparing delegated service with clients' own use of general-purpose AI tests whether delegation reduces their burden. Comparing it with the individual's direct provision of the same service tests whether the agent expands that person's service capacity. Comparing platform configurations while holding the foundation model and the individual's professional resources constant helps identify the contribution of platform mechanisms. Use comparable task objectives, delivery requirements, and deadlines across groups, and document the data, tools, compute, and human assistance in each condition. Prior work has advocated evaluating task performance and cost jointly \cite{ref20}. DelegateEval applies that comparison to the full professional service relationship.

\FloatBarrier
\subsubsection{Client benefits and access to services}\label{sec:6.8.2}
Value to the client includes both the result obtained and the effort and expense required to obtain it. Evaluation should separately record time spent preparing requirements, selecting a service, communicating, waiting, checking results, and handling rework, together with fees paid and final acceptance outcomes. A client who commissions an industry report from an expert's agent, for example, may spend less time finding sources but more time verifying its conclusions. Only by considering all stages together can we determine whether the overall burden falls. Report waiting time separately from active working time. Evaluation should also cover subsequent maintenance included in the service commitment.

For clients with limited domain knowledge, evaluation should also check whether they can detect consequential errors; less time spent checking a result should not automatically be interpreted as greater value. Access to services can be examined through the rate at which suitable services are found, successful commissioning, affordability, and the proportion of clients obtaining results that meet the acceptance criteria. For people previously unable to complete the task themselves or obtain human assistance, report newly available service opportunities and the associated burden. Client satisfaction can provide supplementary evidence, but actual results, acceptance of erroneous outputs, and total effort must still be recorded independently.

\FloatBarrier
\subsubsection{The principal's effort and sustained operation}\label{sec:6.8.3}
An expansion in individual service capacity should be reflected in more accepted services per unit of the principal's time. Count time spent organizing knowledge, carrying out initial configuration, improving the agent over time, conducting expert reviews, approving outgoing communications, taking over in exceptional cases, and providing subsequent maintenance. Report both principal time per accepted service and accepted services per principal-hour, including initial setup investment and performance during steady-state operation. Work performed by other reviewers or platform staff should be recorded separately, so that time saved by the principal can be interpreted alongside any additional work shifted to others. When no principal time is recorded, report service volume and raw inputs, and identify any ratios that are undefined.

For paid services, also record actual receipts, refunds, subsidies, platform fees, and operating expenses. Report net cash income after actual expenditure, with the principal's time listed separately. If the opportunity cost of that time is also monetized, state the rates and calculation method. For free, public-interest, or other non-transactional services, evaluation can focus on time saved, additional opportunities to obtain accepted services, and the investment required to keep them running. Revenue and profitability measures apply only where relevant.

\begin{figure}[!htbp]
\centering
\includegraphics[width=\linewidth,height=0.60\textheight,keepaspectratio]{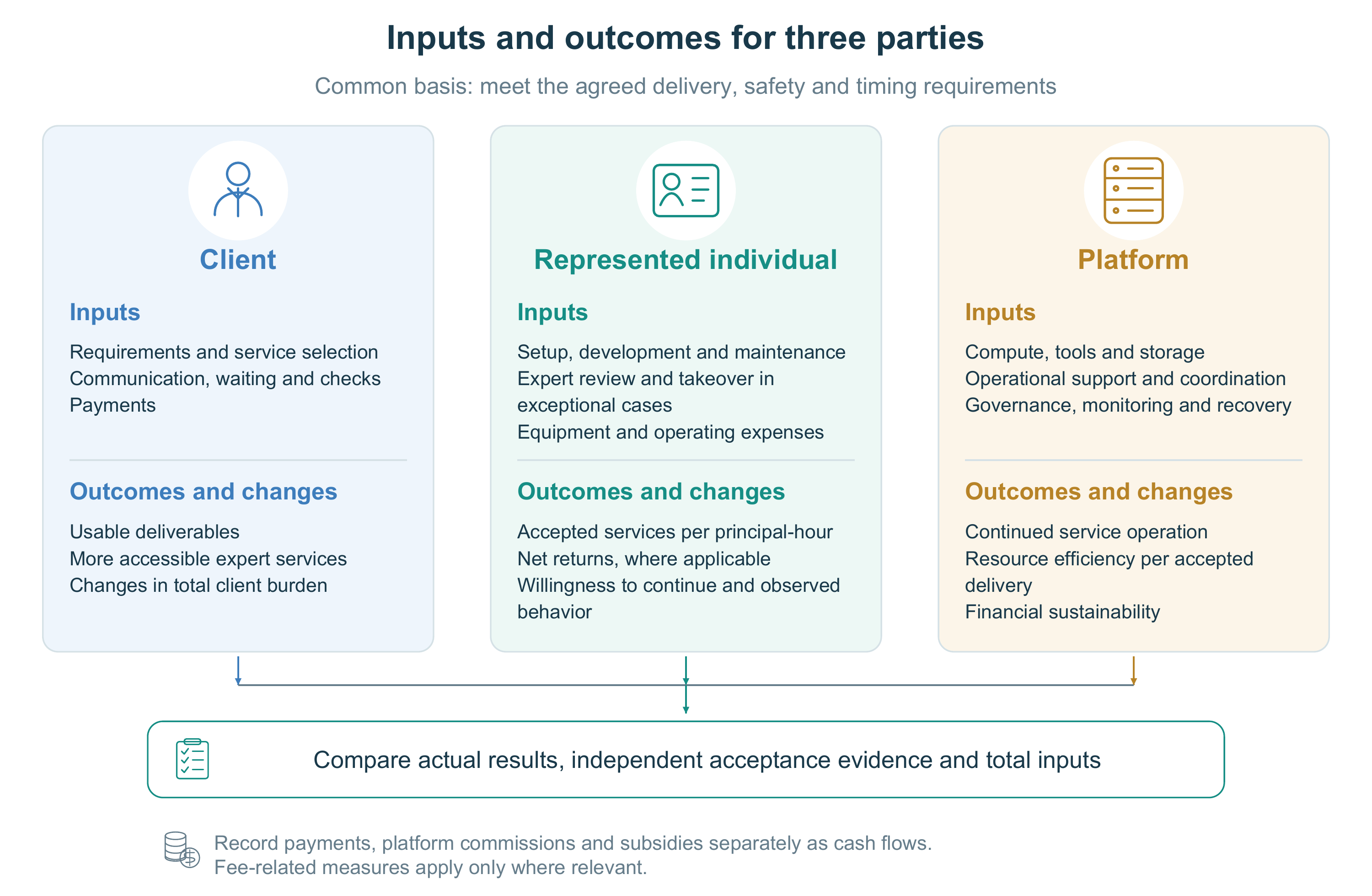}
\caption{Inputs and benefits for clients, represented individuals, and platforms. Under common delivery, safety, and deadline requirements, record each party's investment and what it receives. Payments, platform commissions, and subsidies are tracked separately as financial transfers; fee-related measures apply only in relevant settings. This diagram specifies an evaluation structure, not measured returns.}
\label{fig:7}
\end{figure}

Client feedback, professional reputation, and potential income may encourage individuals to keep developing their agents. This expectation requires evidence from behavior: voluntary maintenance, knowledge updates, time invested, continued operation, and reasons for withdrawal, interpreted alongside service performance and net income. Surveyed intentions can help explain motivation but cannot replace records of sustained effort. The learning and improvement dimension tests capability gains after experience-based updates; this dimension asks whether those gains help individuals continue providing services. Figure~\ref{fig:7} summarizes the relationship between inputs and benefits for the three parties.

\FloatBarrier
\subsubsection{Total cost and platform sustainability}\label{sec:6.8.4}
Account for total cost across initial setup, individual service episodes, and ongoing maintenance. Setup includes organizing materials, preparing examples, configuration, tool integration, and any necessary model adaptation. Costs for a service episode include inference, software calls, human review, rework, and retries after failure. Maintenance includes knowledge and model updates, storage, equipment depreciation, monitoring, and recovery from faults. Record the time of clients, principals, expert reviewers, and platform operators separately. If human time is not monetized, report monetary cost per accepted delivery alongside each party's time investment. Any total-cost figure should specify how labor is valued and which resources are included; retain the original time records as well.

\begin{figure}[!htbp]
\centering
\includegraphics[width=\linewidth,height=0.60\textheight,keepaspectratio]{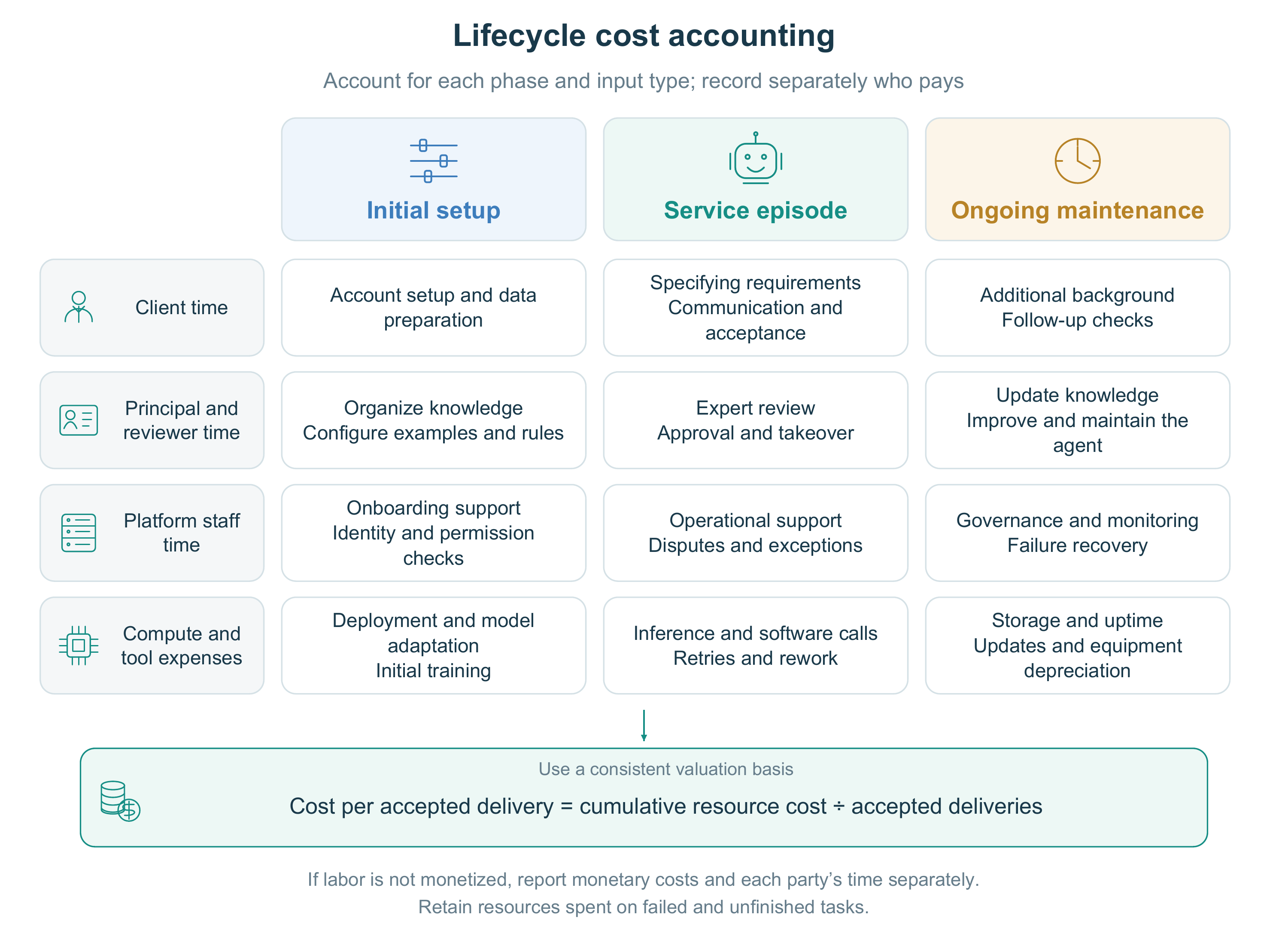}
\caption{Lifecycle cost accounting. Record human time and resource expenditure across setup, service, and maintenance, including resources consumed by failures, retries, and unfinished tasks. Separately identify the hosting arrangement and the party bearing each cost; count each resource cost only once within the stated boundary.}
\label{fig:8}
\end{figure}

Within a predefined observation period, total cost per accepted delivery is cumulative resource cost up to the evaluation endpoint divided by the number of accepted deliveries. Include resources already consumed by failed, canceled, or unfinished tasks in cumulative investment. If there are no accepted deliveries, report that fact and the cumulative investment; unit cost is undefined. State what a short observation window leaves out, including recovery of setup costs and subsequent maintenance obligations. Address these gaps through longer observation or scenario analysis with explicit assumptions. Projected income must not be reported as realized returns. Figure~\ref{fig:8} shows the accounting scope.

Platform hosting and provider self-hosting change how execution resources are configured and who bears their costs. Self-hosting may reduce execution expenses paid directly by the platform while adding equipment, continuous availability, coordination, and recovery costs. Under common quality, safety, and load conditions, report costs borne by the platform and the individual separately, together with total resource cost within the stated accounting boundary. Client payments, platform commissions, and subsidies should be listed separately as financial transfers, and the same expenditure counted only once in aggregate resource cost. Assess platform sustainability using actual revenue, operating expenditure, and dependence on subsidies.

\FloatBarrier
\subsubsection{Scaling services and long-term value}\label{sec:6.8.5}
Scalability must be tested as service volume grows. Vary total request volume, concurrent arrivals, and sustained load separately, and observe accepted deliveries per unit time, the acceptance rate, queues and backlogs, P90/P95 end-to-end latency, the principal's intervention time, and total cost. Define quality, safety, and deadline requirements in advance and keep them comparable across loads. A scalability claim under the tested conditions is supported only if the principal's effort grows more slowly than service volume, accepted services per unit of that time increase, and the specified service requirements continue to be met.

\begin{figure}[!htbp]
\centering
\includegraphics[width=\linewidth,height=0.60\textheight,keepaspectratio]{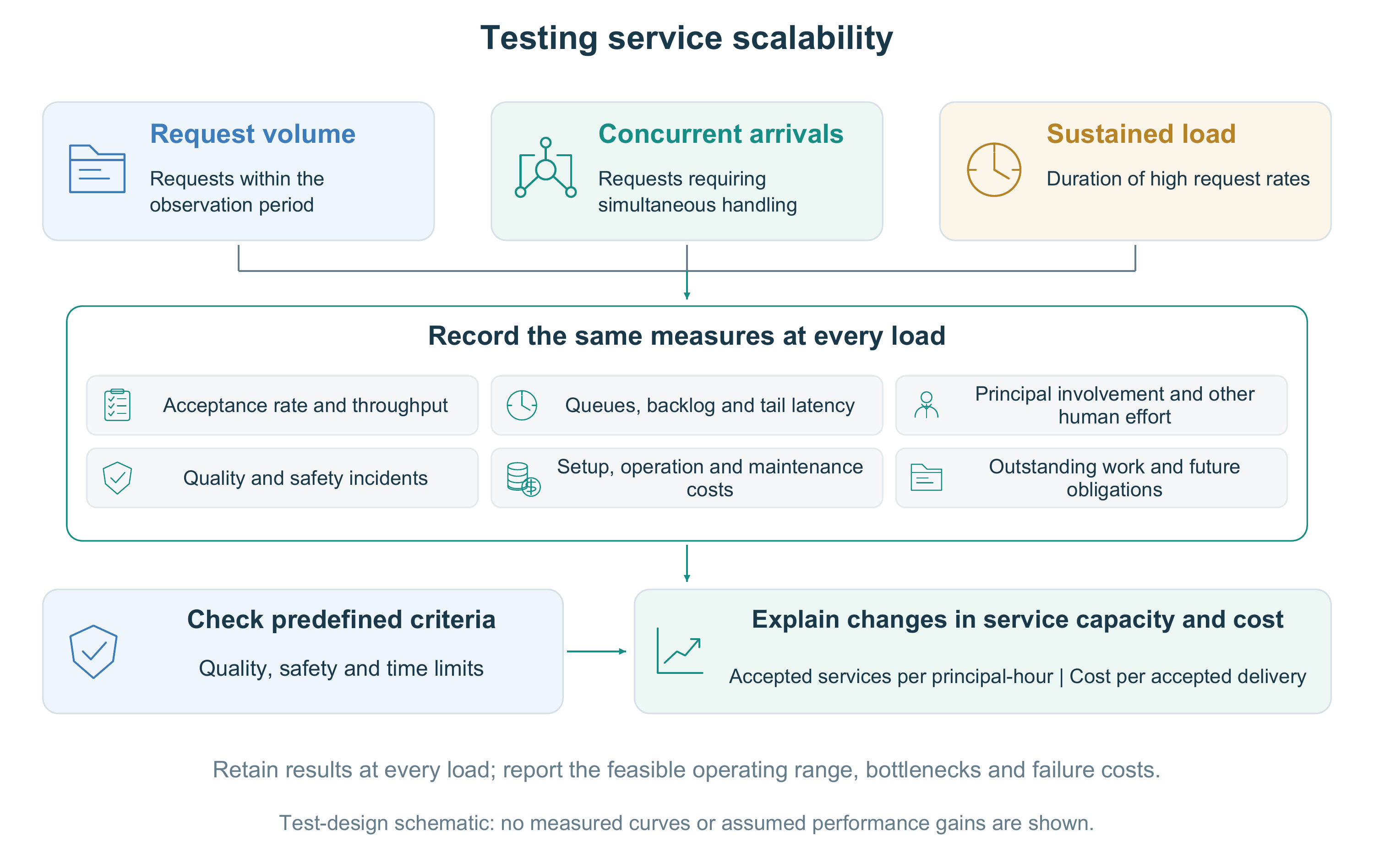}
\caption{Testing service scalability. Increase total volume, concurrency, and sustained load separately. At each load, record service outcomes and all inputs, check the quality, safety, and deadline requirements specified in advance, and then explain changes in capacity and cost. Retain results at every load; the diagram does not assume that growth necessarily improves efficiency.}
\label{fig:9}
\end{figure}

Growth can also expose bottlenecks. A service that requires only occasional approval at low load may develop long queues under greater concurrency, accumulating both client waiting time and maintenance obligations. Retain records of failures, unauthorized actions, and recovery effort at every load, and report the acceptable operating range and the limits encountered first. Reliability research has shown that average task performance does not adequately describe consistency, robustness, or the severity of failures \cite{ref21}. Magentic Marketplace also found that providing more search results reduced consumer welfare in its simulated markets \cite{ref11}. Scaling tests must therefore observe service supply, client choice, and actual fulfillment together. Figure~\ref{fig:9} outlines this evaluation process.

Long-term value should also be visible in repeat engagements. Compare the effort required for background explanation, requirements clarification, verification, rework, and maintenance in initial and subsequent commissions to determine whether existing information and service experience reduce the burden of collaboration. Continuity and learning effects are verified in their respective dimensions; this dimension records the resulting changes in time, cost, and service capacity. At the end of the observation period, list outstanding maintenance commitments, refunds, and disputes so that interim returns can be interpreted alongside continuing obligations.

\FloatBarrier
\subsubsection{Distribution of benefits and reporting}\label{sec:6.8.6}
To evaluate broad access to AI-enabled services, examine who benefits. Group observations by factors such as professional background, AI proficiency, initial data and equipment, and clients' domain experience. Retain failed creation attempts, services that could not be published, and unfinished commissions. A platform may deliver substantial benefits to well-resourced experts while people with limited AI knowledge or equipment still struggle to keep their services running. Combine these observations with the creation outcomes in Section~\ref{sec:6.1}, reporting access to accepted services, time and financial burdens, provider net income, and service withdrawals for each group.

Report delivery quality, client burden, principal effort, resource cost, and service scale together. For each result, state the observation period, unit of analysis, denominator, comparison conditions, and human assistance. Report variation across repeated runs and appropriate uncertainty estimates. Table~\ref{tab:6} summarizes the main measures and underlying records to retain. Where conditions differ or evidence is insufficient, state the conditions to which the conclusions apply. Long-term operational and economic outcomes that have not yet been measured remain questions for subsequent testing.

\begingroup\small
\arrayrulecolor{TableGrid}\setlength{\arrayrulewidth}{0.4pt}
\setlength{\tabcolsep}{6pt}\setlength{\extrarowheight}{3pt}\renewcommand{\arraystretch}{1.12}
\begin{longtable}{|>{\raggedright\arraybackslash}p{\dimexpr 0.22\linewidth-2\tabcolsep-1.5\arrayrulewidth\relax}|>{\raggedright\arraybackslash}p{\dimexpr 0.78\linewidth-2\tabcolsep-1.5\arrayrulewidth\relax}|}
\caption{Key measures and underlying records for efficiency and value. Report each result separately; fee and profitability measures apply only in relevant settings.}\label{tab:6}\\
\hline
\rowcolor{TableHead} \textbf{Perspective} & \textbf{Key measures and records to retain} \\ \hline
\endfirsthead
\multicolumn{2}{l}{\small Table 6 (continued)}\\
\hline
\rowcolor{TableHead} \textbf{Perspective} & \textbf{Key measures and records to retain} \\ \hline
\endhead
\endfoot
Client & Accepted outputs; opportunities to obtain services; preparation and communication time; waiting and verification time; rework; rate at which clients accept erroneous outputs; fees paid \\ \hline
\rowcolor{TablePale} Represented individual & Time spent on setup, development, review, intervention, and maintenance; accepted services per principal-hour; net income where applicable \\ \hline
Platform and resources & Setup, operating, maintenance, failure, and recovery costs; costs borne by each party; platform revenue and dependence on subsidies \\ \hline
\rowcolor{TablePale} Scale and long-term operation & Throughput of accepted deliveries and acceptance rate, tail latency, backlog, risk, and outstanding obligations at different loads; burden of repeat engagements \\ \hline
Population groups & Creation and continued service operation by group; opportunities to obtain accepted services; time and financial burdens; benefits and reasons for withdrawal \\ \hline
\end{longtable}
\endgroup

\FloatBarrier
\section{Limitations and future work}\label{sec:7}
DelegateEval is currently an evaluation framework, not a complete, executable benchmark. This paper defines the systems under study and proposes eight evaluation dimensions, task categories, acceptance principles, and experimental designs. A complete dataset, an executable environment, a common scoring protocol, and results across foundation models are still missing. The task taxonomy, proposed test cases, and diagrams illustrate the evaluation design; they are neither a standalone test suite nor empirical evidence. The immediate priority is to build these components so that other researchers can run, verify, and compare results under explicit conditions.

\FloatBarrier
\subsection{Current limitations}\label{sec:7.1}
The task categories and proposed test cases do not yet constitute a complete dataset. The work does not yet provide a full, releasable set of inputs, configurations of individuals' knowledge and methods, client requests, event sequences, reference outputs, or independent acceptance materials. Coverage, sample size, and training, development, and test splits also remain to be defined. Dataset construction must resolve how to organize samples across individuals, clients, and time; retain creation failures; and prevent leakage through shared materials or histories.

The acceptance principles have not yet been implemented as a common scoring and reporting protocol. The paper distinguishes diagnostic scores, overall acceptance, minimum safety requirements, and insufficient evidence, but executable scoring scripts, review rubrics, and complete statistical rules are still needed. Units of analysis, denominators, timeout and failure handling, missing evidence and inapplicable items, repeated runs, weights, and thresholds must all be specified in advance. Reviewers of open-ended tasks also need to calibrate their judgments for consistency. The common protocol should establish a comparable basis while retaining industry-specific acceptance requirements and preventing the same result from being counted more than once across tasks or dimensions.

Baselines and controlled comparisons across foundation models are also missing. Models from different families and of different sizes have not yet been compared on the same dataset, environment, and protocol, and there is insufficient evidence to separate the contributions of the foundation model, personal resources, platform mechanisms, and human support. Consequently, we cannot yet establish whether conclusions about a platform hold when the foundation model changes. An adaptation gain on one model is also insufficient to establish an advantage for the service as a whole. The effects of individual knowledge adaptation, memory, skills, and workflow mechanisms should be tested through controlled comparisons and ablations with clearly specified resources.

These gaps limit the empirical scope of the present conclusions. Suitability for real creators, validity of the review rubrics, long-term continuity, learning gains, scalability, economic sustainability, and governance performance all require validation. They remain questions for subsequent testing. The current framework does not support platform rankings or claims that high leverage or profitability has been demonstrated. An absence of observed safety incidents, by itself, also does not establish long-term safety.

\FloatBarrier
\subsection{Next steps}\label{sec:7.2}
First, build an initial dataset with complete materials and explicit coverage. Prioritize tasks such as code modification, research reports, document maintenance, and tutoring. Develop the proposed test cases into complete tasks containing individual configurations, client requests, authorized materials, event timelines, reference outputs, and independent acceptance materials. Organize data splits by individual, project, and shared history, check for near-duplicate tasks and data leakage, and record professional and technical backgrounds, completeness of the materials, and creation failures. The first release can be modest in scale, but coverage of each dimension should be clear so that it provides a basis for broader tasks and studies with real participants.

Second, deliver an evaluation environment that others can run and verify. Implement an evaluation process covering creation, individual service episodes, and replay of longitudinal service records. Fix tool and dependency versions, and support state resets, event injection, authorization changes, fault recovery, freezing and archiving deliverables, and log export. Keep evaluation-side validators isolated from the system under test. Test them using acceptable outputs, cases with clear defects, reasonable alternatives, and samples with incomplete evidence. Publish installation instructions, configuration examples, and end-to-end demonstrations so that independent users can reproduce acceptance checks from the original materials and distinguish environment faults from failures of the system under test.

Next, invite creators and clients from varied backgrounds to pilot studies. Calibrate the scoring criteria and reviewers' judgments, then fix task endpoints, resource and human-assistance budgets, scoring rules, acceptance criteria, and exception handling before formal testing. Under the same protocol, compare foundation-model families and sizes, along with configurations using individual knowledge, model adaptation, and workflow improvements. Test whether they improve performance on unseen tasks while retaining existing capabilities, and record the associated costs. Report results for the eight dimensions, accepted deliveries, safety incidents, and human effort separately; use repeated runs to assess stability; and release the necessary versions, configurations, and logs so that contributions from foundation models, individual expertise, platform mechanisms, and additional human work can be distinguished.

The first benchmark release is complete only when independent researchers can run the evaluation using its data, environment, and protocol and verify the cross-model results.

Building on that release, proceed to 30-day and 90-day observations of real services and tests under different loads, examining long-term learning, authorization changes, service scale, and economic value. Report simulated business time separately from elapsed operating time. Release a new version when the data or protocol changes materially, preserving previous results for later comparison.

\FloatBarrier

\end{document}